\pdfoutput=1
\documentclass{article}
\usepackage{natbib}
\usepackage[colorlinks=true, citecolor=blue, urlcolor=black]{hyperref}
\usepackage{graphicx}
\usepackage{amssymb}
\usepackage{amsmath}
\usepackage{color,soul,cancel,subfigure}
\usepackage{setspace,mathrsfs,soul,cancel}
\usepackage{xargs} 
\usepackage{dsfont}

\def\bSig\mathbf{\Sigma}

\newcommand{\iid}{i.i.d.}%\text{i.i.d.}

\begin{document}

% Title of paper
\title{Sequential Dirichlet Process Mixtures of Multivariate Skew $t$-distributions for \\Model-based Clustering of Flow Cytometry Data}

% List of authors, with corresponding author marked by asterisk
\author{Boris P. Hejblum$^{1,2,*}$
Chariff Alkhassim$^{1,2}$
Raphael Gottardo$^{3}$, \smallskip\\
Fran\c cois Caron$^{4}$, and
Rodolphe Thi\'ebaut$^{1,2}$\medskip
\\
\small$^{1}$Univ. Bordeaux, ISPED, Bordeaux Population Health Research Center, \qquad\qquad\,\\\small Inserm U1219, Inria SISTM, 33000 Bordeaux, France\qquad\qquad\qquad\qquad\qquad\quad~\\
\small$^{2}$Vaccine Research Institute (VRI), 94010 Cr\'eteil, France\qquad\qquad\qquad\qquad\qquad~\,\\
\small$^{3}$Fred Hutchinson Cancer Research Center, Seattle, Washington, U.S.A.\qquad\qquad~\,\,\,\,\\
\small$^{4}$Department of Statistics, University of Oxford, Oxford, U.K.\hfill\medskip \\
\small$^*$\textit{boris.hejblum@u-bordeaux.fr}\qquad\qquad\qquad\qquad\qquad\qquad\qquad\qquad\qquad\qquad\qquad\qquad\hfill
%$^{\dagger}$ equally contributed
}

% Running headers of paper:
\markboth%
% First field is the short list of authors
{BP Hejblum, C Alkhassim, R Gottardo, F Caron, R Thiébaut}
% Second field is the short title of the paper
{Sequential Dirichlet Process Mixtures of Multivariate Skew $t$-distributions}

\maketitle

% Add a footnote for the corresponding author if one has been
% identified in the author list

\begin{abstract}
{Flow cytometry is a high-throughput technology used to quantify multiple surface and intracellular markers at the level of a single cell. This enables to identify cell sub-types, and to determine their relative proportions. Improvements of this technology allow to describe millions of individual cells from a blood sample using multiple markers. This results in high-dimensional datasets, whose manual analysis is highly time-consuming and poorly reproducible. While several methods have been developed to perform automatic recognition of cell populations, most of them treat and analyze each sample independently. However, in practice, individual samples are rarely independent (e.g. longitudinal studies). Here, we propose to use a Bayesian nonparametric approach with Dirichlet process mixture (DPM) of multivariate skew $t$-distributions to perform model based clustering of flow-cytometry data. DPM models directly estimate the number of cell populations from the data, avoiding model selection issues, and skew $t$-distributions provides robustness to outliers and non-elliptical shape of cell populations. To accommodate repeated measurements, we propose a sequential strategy relying on a parametric approximation of the posterior. We illustrate the good performance of our method on simulated data, on an experimental benchmark dataset, and on new longitudinal data from the DALIA-1 trial which evaluates a therapeutic vaccine against HIV. On the benchmark dataset, the sequential strategy outperforms all other methods evaluated, and similarly, leads to improved performance on the DALIA-1 data. We have made the method available for the community in the R package \texttt{NPflow}.
}\bigskip\medskip\bigskip\smallskip
\end{abstract}

\section{Introduction}
\label{s:intro}

Flow cytometry is a high-throughput technology used to quantify multiple surface and intracellular markers at the level of single cell. More specifically, cells are stained with multiple fluorescently-conjugated monoclonal antibodies directed to cell surface receptors (such as CD4) or intracellular markers (such as cytokines) to determine the type of cell, their differentiation and their functionality. With the improvement of this technology leading currently to the measurement of up to 18 at the same time (using 18 colors for Flow cytometry), multi-parametric description of millions of individual cells can be generated. 

Analysis of such data is generally performed manually. This results in analyses that are: i) poorly reproducible \citep{Aghaeepour2013}, ii) expensive (highly time-consuming) and iii) as a result of ii), focused on specific cell populations (i.e. specific combination of markers), ignoring other cell populations. There has been an effort in the recent years to offer automated solutions to overcome these limitations \citep{Lo2008, Aghaeepour2013, Gondois-Rey2016}. Quite a lot of different methodological approaches have been proposed to perform automatic recognition of cell populations from flow cytometry data. Clustering methods related to the k-means were proposed, such as L2kmeans \citep{Aghaeepour2013}, flowMeans \citep{Aghaeepour2011}. Model based clustering methods relying on finite mixture models such as flowCust/merge \citep{Lo2008, Finak2009}, FLAME \citep{Pyne2009}, SWIFT \citep{Naim2014} were also proposed, as well as dimension reduction methods such as MM and MMPCA \citep{Sugar2010}, SamSPECTRAL \citep{Zare2010}, FLOCK \citep{Qian2010}. All those approaches requires the number of cell populations to be fixed in advance, and resort to various criteria to determine the number of cell populations. Finally,  several authors \citep{Chan2008, Lin2013, Cron2013, Dundar2014}, proposed nonparametric Bayesian mixture models of Gaussian distributions, that directly estimate the number of cell populations. All these methods, except those of \cite{Lin2013}, of \cite{Cron2013} and of \cite{Dundar2014}, were evaluated by \cite{Aghaeepour2013}.

However, there is still room for improvement, especially in the estimation of the suitable number of cell populations as well as in the identification of rare cell populations. In addition, most of those previous approaches have been proposed for single sample analysis, except for \cite{Cron2013} who proposed to use hierarchical Dirichlet process mixture (DPM) of Gaussian distribution models to analyze multiple samples simultaneously. Yet in the case of repeated measurements of flow cytometry data, it can be useful to perform analysis as the samples are acquired (samples are often collected across several time points in a population of patients). In such a case, one would want to use previously acquired sample as informative prior information in the analysis of a new sample. In this paper, the proposed approach includes a strategy of sequential approximations of the posterior distribution for multiple data samples, presented in Section \ref{seqpriors}. Our approach offers three advantages: i) it quantifies the uncertainty around the posterior clustering estimate, ii) it can make use prior knowledge to inform on the structure of the data, potentially building up on previous analyses, and iii) it allows the analysis of multiple samples without requiring to process all the data at once, alleviating both the computational burden and the necessity for all data to be readily available before any analysis can be performed.

The automatic recognition of cell populations from flow cytometry data is a difficult task which can be seen as an unsupervised clustering problem \citep{Lo2008}. It is characterized by two big challenges. First, the total number of cell populations to identify is unknown. Second, the empirical distributions of the populations are heavily skewed, even when optimal transformation of the data is applied \citep{Lo2008, Pyne2009, Lo2012}, and the data generally present many outliers. To address all these points together, our approach consider a Bayesian nonparametric model-based approach, where the flow cytometry data are assumed to be drawn from a DPM skew-$t$ distributions. First, this approach enables the number of cell populations to be inferred from the data, and avoids the challenging problem of model selection. Second, it has been demonstrated that the Gaussian assumption for the parametric shape of a cell population fits poorly flow cytometry data \citep{Mosmann2014}. Indeed, even after state-of-the-art transformation of raw cytometry data, such as the biexponential transformation \citep{Finak2010}, cell population distributions are typically skewed. \cite{Pyne2009} have showed the advantages of the skew $t$-distribution \citep{Azzalini2003} for modeling cell subpopulations in flow cytometry data. The skew $t$-distribution is a generalization of the skewed normal distribution, with a heavier tail which makes it more robust to outliers. \cite{Fruhwirth-Schnatter2010a} proposed a finite mixture model of skew $t$-distributions. We extend this model to the infinite mixture case in a Bayesian nonparametric framework. Of interest, quantifying the uncertainty around the estimated partition is straightforward in this Bayesian paradigm, from the posterior distribution of the partition. While a skewed distribution could be fit either by a skew-t or a mixture of Gaussians, using the latter requires to separate the estimation of the overall number of clusters from the skewness, while the proposed approach jointly estimate those two and thus takes into account the uncertainty associated with both. Furthermore, the use of a Bayesian framework enables the use of informative priors. In the case of repeated measurements for instance, we propose to sequentially estimate the posterior partition of flow cytometry using posterior information from time point $t$ as prior information for time point $t+1$.

The proposed approach is applied to simulated data, to a benchmark clinical dataset from \cite{Aghaeepour2013}, and to an original experimental dataset from a phase I HIV clinical trial DALIA-1. The method is implemented in the R package NPflow, available on the CRAN at \url{https://CRAN.R-project.org/package=NPflow}.

\section{Statistical Model}
\label{s:methods}

\subsection{Problem set-up}
\label{ss:model}

In this Section we first consider that we have only one sample per subject. The case of the sequential estimation of multiple datasets will be addressed in Section \ref{seqpriors}. We consider that we have data $\boldsymbol{y}_{c}\in\mathds{R}^{d}$,
$c=1,\ldots,C$  corresponding to the vector of fluorescence intensities measured for the cell $c$. Typically, the observations $\boldsymbol{y}_c$ have been transformed (to help visualization and gating) from the raw measurements of fluorescence through a biexponential or Box-Cox transformation \citep{Finak2010}. We assume that these observations are independent and identically distributed (\iid) from some unknown distribution\ $F$:%
\begin{equation}
\boldsymbol{y}_{c}|G\overset{\tiny{\iid}}{\sim}F\text{ for }c=1\ldots,C
\label{eq:yF}
\end{equation}
where $F$ is a mixture of distributions:%
\begin{equation}
F(\boldsymbol{y})=\int_{\boldsymbol{\Theta}}f_{\boldsymbol{\theta}}(\boldsymbol{y})G(d\boldsymbol{\theta}) \label{eq:mixture}%
\end{equation}
with $f_{\theta}(\boldsymbol{y})$ a known probability density function, parameterized by $\boldsymbol{\theta}\in\Theta$, a set of parameters, and defining the shape of a cluster. $G$ is the unknown mixing distribution, which carries the weights and locations of the mixture components. In a parametric approach, $G=\sum_{k=1}^K\pi_k\delta_{\boldsymbol{\theta}_k}$ where $\pi_k$ is the weight of the $k^\text{th}$ mixture component. Maximum likelihood or Bayesian estimates of $F$ can be derived for such models \citep{Biernacki2000}. In a nonparametric perspective (where the number of clusters is unknown) $G$ is written as an infinite sum of atoms: $G=\sum_{k=1}^{+\infty}\pi_k\delta_{\boldsymbol{\theta}_k}$. The Dirichlet process is a conjugate prior for the infinite atomic discrete distribution, which makes it very useful for unsupervised clustering approaches.

\subsection{Dirichlet process mixture model}
We assume that the random mixing distribution $G$ is drawn from a Dirichlet process \citep{Ferguson1973}:
\begin{equation}
G\sim \text{DP}(\alpha,G_0)
\label{eq:DP}
\end{equation}
where $\text{DP}(\alpha,G_{0})$ denotes the Dirichlet process of scale parameter $\alpha>0$ and base probability distribution $G_{0}$. A draw $G\sim\text{DP}(\alpha,G_{0})$ is almost surely discrete and takes the following form \citep{Sethuraman1994}:
\begin{equation}
G=\sum_{k=1}^{+\infty}\pi_{k}\delta_{\boldsymbol{\theta}_k}%
\end{equation}
where the $\theta_{k}$ are \iid~from the base distribution $G_{0}$ and independent of the weights, $\boldsymbol{\pi}=(\pi_{k})_{k=1,2,\ldots}$, which are drawn from a so-called ``stick-breaking" distribution: $$\pi_{k}=\beta_{k}\prod_{j=1}^{k-1}(1-\beta_j)$$
with $\beta_{k}\overset{\tiny{\iid}}{\sim}\text{Beta}(1,\alpha)$ for $k=1,2,\ldots$\,\,. We write $\boldsymbol{\pi}\sim \text{GEM}(\alpha)$ the Griffiths-Engen-McCloskey (GEM) distribution \citep{Pitman2006}.  The model defined by Equations~\eqref{eq:yF},~\eqref{eq:mixture} and ~\eqref{eq:DP} yields the following hierarchical model known as a Dirichlet process mixture model \citep{Lo1984, Escobar1995, Teh2010} with a Gamma hyperprior on the concentration parameter $\alpha$:
%\begin{subequations}
%\begin{align}
%	G\,\big|\,\alpha, G_0	&\sim\text{DP}(\alpha,G_{0})\\
%\intertext{and for $c=1,\ldots,C$}
%	\boldsymbol{\theta}_{c}\,\big|\,G  	&\sim G\\
%	y_{c}\,\big|\,\boldsymbol{\theta}_{c}&\sim f_{\boldsymbol{\theta}_{c}}
%\end{align}
%\label{eq:DPM}
%\end{subequations}

%Because a lot of $\boldsymbol{\theta}_c$ are duplicates, it is useful at this point to introduce latent variables $\ell_{c}$ for $c=1,\ldots,C$, giving the latent parameters of the mixture component from which $y_{c}$ is drawn. The hierarchical model can then be re-written as follows:

\begin{subequations}
\begin{align}
	\alpha | a,b  &\sim \text{Gamma}(a, b)\\
	\boldsymbol{\pi}\,\big|\,\alpha 					&\sim\text{GEM}(\alpha)\\
\intertext{for $k=1,2,\ldots$}
\boldsymbol{\theta}_{k}\,\big|\,G_0    			&\sim G_0\\
\intertext{for $c=1,2,\ldots,C$}
	\ell_c\,\big|\,\boldsymbol{\pi} 					&\sim \text{Mult}(\boldsymbol{\pi})\\	
	y_{c}\,\big|\,\ell_c , (\boldsymbol{\theta}_k) &\sim f_{\boldsymbol{\theta}_{\ell_c}}
\end{align}
\label{eq:DPM}
\end{subequations}
where $\ell_c$ is an allocation variable indicating to which cluster is associated cell $c$.

The base distribution\ $G_0$ tunes the prior information we have about the cluster locations. The parameter $\alpha$ tunes the prior distribution on the overall number of clusters $K$ that will be discovered within $C$ data. In particular we have $\mathds{E}[K] =\sum_{c=0}^{C-1}\frac{\alpha}{\alpha+c}$.

\subsection{Multivariate skew-t distribution}
We now consider the choice of the parametric density $f_\theta$ which is a skew-t distribution.
\subsubsection{Skew-normal distribution}
\cite{Fruhwirth-Schnatter2010a} present a parametrization of the multivariate skew Normal distribution defined by \cite{Azzalini1996} which leads to the following probability density function:
\begin{equation}
f_{\mathcal{SN}}(\boldsymbol{y}; \boldsymbol{\xi}, \boldsymbol{\Omega}, \boldsymbol{\eta}) = 2\phi(\boldsymbol{y}-\boldsymbol{\xi}; \boldsymbol{\Omega})\Phi(\boldsymbol{\eta}'\boldsymbol{\omega}^{-1}(\boldsymbol{y}-\boldsymbol{\xi}))
\end{equation}
with $\phi(\cdot; \boldsymbol{\Omega})$ the probability density function of the multivariate Normal distribution with zero mean $\mathcal{N}(0,\Omega)$ and $\Phi(\cdot)$ the cumulative density function of the standard univariate Normal distribution $\mathcal{N}(0,1)$.

\cite{Fruhwirth-Schnatter2010a} propose a random-effects model representation of such a skew Normal distribution, with truncated normal random effects:
\begin{equation}
\boldsymbol{Y}= \boldsymbol{\xi} + \boldsymbol{\psi}Z + \boldsymbol{\varepsilon}
\label{eq:skewnormal}
\end{equation}
with $Z\sim\mathcal{N}_{[0;+\infty[}(0,1)$ a truncated univariate standard Normal distribution and $\boldsymbol{\varepsilon}\sim\mathcal{N}(\boldsymbol{0}, \boldsymbol{\Sigma})$ a multivariate Normal distibution with zero mean. The original parameters can be recovered from:
\begin{equation}
\boldsymbol{\Omega}=\boldsymbol{\Sigma} + \boldsymbol{\psi}\boldsymbol{\psi}',\quad \boldsymbol{\eta} = \frac{1}{\sqrt{1-\boldsymbol{\psi}'\boldsymbol{\Omega}^{-1}\boldsymbol{\psi}}}\boldsymbol{\omega}\boldsymbol{\Omega}^{-1}\boldsymbol{\psi}
\end{equation}
\subsubsection{The skew t-distribution}

Let $\boldsymbol{X}\sim\mathcal{SN}(\boldsymbol{0}, \boldsymbol{\Omega}, \eta)$ and $W\sim\text{Gamma}(\frac{\nu}{2}, \frac{\nu}{2})$. If $\boldsymbol{Y}$ has the following stochastic representation:
\begin{equation}
\boldsymbol{Y} = \boldsymbol{\xi} + \frac{1}{\sqrt{W}}\boldsymbol{X}
\label{eq:skewt}
\end{equation}
then it follows a multivariate skew $t$-distribution $\boldsymbol{Y}\sim\mathcal{ST}(\boldsymbol{\xi}, \boldsymbol{\Omega}, \boldsymbol{\eta}, \nu)$ \citep{Azzalini2003}. Equation \eqref{eq:skewt} can be expressed as the following random effect model
\begin{equation}
\boldsymbol{Y} = \boldsymbol{\xi} + \boldsymbol{\psi}\frac{Z}{\sqrt{W}}+ \frac{\boldsymbol{\epsilon}}{\sqrt{W}}
\label{eq:skewt2}
\end{equation}

\begin{figure}[!h]
	\centerline{\includegraphics[width=0.85\textwidth]{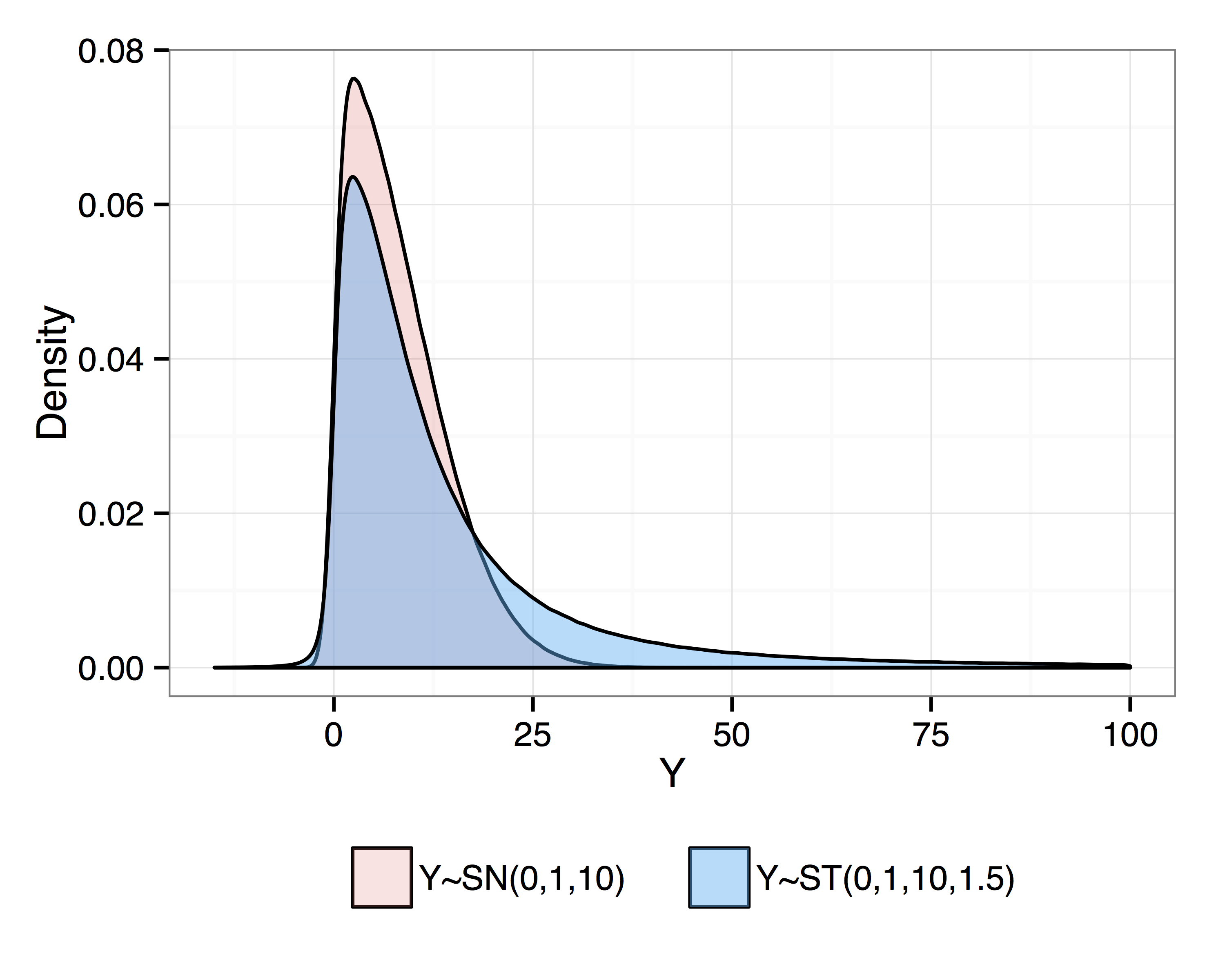}}
	\caption{Density probability function of univariate skew Normal $\mathcal{SN}(\xi=0,\psi=10, \sigma=1)$ and skew t $\mathcal{ST}(\xi=0,\psi=10,\sigma=1,\nu=1.5)$ distributions}
	\label{fig:ExSNST}
\end{figure}

Following the same parametrization as \cite{Fruhwirth-Schnatter2010a}, we write the density of a multivariate skew t-distibution as:
\begin{align}
f_{\mathcal{ST}}(\boldsymbol{y}; \boldsymbol{\xi}, \boldsymbol{\Omega}, \boldsymbol{\eta}, \nu) = & 2f_\mathcal{T}(\boldsymbol{y}; \boldsymbol{\xi}, \boldsymbol{\Omega}, \nu)\\
& \times T_{\nu + d}\left(\boldsymbol{\eta}'\boldsymbol{\omega}^{-1}(\boldsymbol{y}-\boldsymbol{\xi})\sqrt{\frac{\nu + d}{\nu + Q_y}}\right)\notag
\end{align}
with $\boldsymbol{\omega} = \sqrt{Diag(\boldsymbol{\Omega})}$, $Q_y=(\boldsymbol{y}-\boldsymbol{\xi})'\boldsymbol{\Omega}^{-1}(\boldsymbol{y}-\boldsymbol{\xi})$, $f_\mathcal{T}$ the multivariate Student $t$-distribution probability density function, and $T_\nu$ the cumulative distribution function of the scalar standard Student $t$-distribution with $\nu$ degrees of freedom. Figure \ref{fig:ExSNST} shows an example of such distributions, highlighting the skewness of both the skew Normal and the skew t and the heavier tail of the skew t distribution.

\subsection{Dirichlet process mixture of skew $t$-distribution}

Let $G_0$ be the base distribution of a Dirichlet process in a DPM combining model \eqref{eq:DPM} with a random-effects model representation \eqref{eq:skewt2} of the skew $t$-distribution. $G_0$ is the product of a structured Normal inverse Wishart ($sNiW$) and of a prior on $\nu$, the degree of freedom of the skew-t: $G_0=sNiW(\xi_0, \psi_0, B_0, \Lambda_0, \lambda_0)P_{0,\nu}$.
Our proposed model is fully written as follows:

\begin{subequations}
\begin{align}
	\alpha | a,b  &\sim \text{Gamma}(a, b)\\
	\boldsymbol{\pi}\,\big|\,\alpha  &\sim \text{GEM}(\alpha)\\
\intertext{for $k=1,2,\ldots$}
\boldsymbol{\xi}_k, \boldsymbol{\psi}_k, \boldsymbol{\Sigma}_k, \nu_k\ &\sim G_0\\	
\intertext{for $c=1,2,\ldots,C$}	
\ell_c\,\big|\,\boldsymbol{\pi}&\sim \text{Mult}(\boldsymbol{\pi})\\
	\gamma_c \,\big|\, \ell_c, (\nu_k)&\sim\text{Gamma}\left(\frac{\nu_{\ell_c}}{2}, \frac{\nu_{\ell_c}}{2}\right)\\
	s_c\,\big|\, \gamma_c &\sim \mathcal{N}_{[0,+\infty[}\left(0,\frac{1}{\gamma_c}\right)\\
		\boldsymbol{y}_{c}\,\big|\,\ell_c, \gamma_c, s_c, (\boldsymbol{\xi}_k, \boldsymbol{\psi}_k, \boldsymbol{\Sigma}_k) &\sim\mathcal{N}\left( \boldsymbol{\xi}_{\ell_c} + \boldsymbol{\psi}_{\ell_c}s_c,\frac{1}{\gamma_c}\boldsymbol{\Sigma}_{\ell_c}\right)
\end{align}
\label{modelFinal}
\end{subequations}

\subsection{Discussion on the model assumptions}
\label{priors}

In model \eqref{modelFinal}, the base distribution parameter $G_0$ conveys the prior information on the cluster parametric shape. For the parameters $\boldsymbol{\xi_k}$, $\boldsymbol{\psi_k}$ and $\boldsymbol{\Sigma_k}$, we have conditional conjugacy with the random-effects model representation using joint priors taking the form of a structured Normal-inverse-Wishart distribution. See Appendix \ref{app:GibbsSamp} for details. \cite{Fruhwirth-Schnatter2010a} pointed out that the prior on $\boldsymbol{\Sigma}_k$ can have a big impact on the posterior number of clusters. Indeed, setting the scale of the prior on $\boldsymbol{\Sigma}_k$ too small will result in an inflated number of clusters in the posterior, whereas too large values tend to cluster all the observations together. Adding a Wishart hyperprior on $\boldsymbol{\Sigma}_k$, that carries on conjugacy with the inverse-Wishart, enables us to reduce this impact of the prior \citep{Fruhwirth-Schnatter2010a, Huang2013a}. Assuming prior independence between each $\nu_k$ and also from the three parameters mentioned above, we can use any of the three priors proposed in \cite{Juarez2010} for instance (such as an objective Jeffrey's prior, see Appendix \ref{app:GibbsSamp}).

\section{Estimation}
\label{estim}

\subsection{Posterior Estimation via Gibbs sampling}
For making inference on the model \eqref{modelFinal}, MCMC methods can be used to sample the partition $\{\ell_{1:C}\}$ and the corresponding cluster parameters $\{\theta^*_{k}\}=\left\{\{\boldsymbol{\xi}^*_k\}, \{\boldsymbol{\psi}^*_k\}, \{\boldsymbol{\Sigma}^*_k\}, \{\nu^*_k\}\right\}$ from the marginal posterior distribution. Extending results from \cite{Fruhwirth-Schnatter2010a} and \cite{Caron2014}, it is possible to implement an efficient and valid partially collapsed Gibbs sampler with a Metropolis-Hastings step \citep{VanDyk2008, VanDyk2015}. The use of slice sampling \citep{Neal2003, Kalli2011} enables the straightforward parallelization of the latent allocation sampling (thanks to conditional conjugacy) in such an MCMC algorithm (even in the skew-normal and skew-$t$ cases), which can lead to substantial computation speed up when the number of observations $C$ (cells) per sample increases. Each iteration of our Gibbs sampler proceeds in the following order (details are provided Appendix \ref{app:GibbsSamp}):

\begin{enumerate}
	\item Update the concentration parameter $\alpha$ given the previous partition $\{\ell_{1:C}\}$ using the data augmentation technique from \cite{Escobar1995}.

	\item Update the mixing distribution $G$ given $\alpha$, $\{\boldsymbol{\xi}_{k}\}$, $\{\boldsymbol{\psi}_{k}\}$, $\{\boldsymbol{\Sigma}_{k}\}$ and the base distribution $G_0$ via slice sampling.

\item For $c=1,\dots,C$  update the individual skew parameter $s_c$ given $\{\boldsymbol{\xi}_{k}\}$, $\{\boldsymbol{\psi}_{k}\}$, $\{\boldsymbol{\Sigma}_{k}\}$ and the new $\ell_c$.

\item Update $\{\boldsymbol{\xi}_{k}\}$, $\{\boldsymbol{\psi}_{k}\}$, $\{\boldsymbol{\Sigma}_{k}\}$ given the base distribution $G_0$, the updated partition $\{\ell_{1:C}\}$ and the updated individual skew parameters $\{s_{1:C}\}$.

\item Finally jointly update the degrees of freedom and  the individual scale factors $(\{\nu_{k}\}, \{\gamma_{1:C}\})$ in an Metropolis-Hastings (M-H) within Gibbs step. First an M-H step is performed to update the $\{\nu_{k}\}$ where the $\{\gamma_{1:C}\}$ are integrated out, immediately followed by a Gibbs step to sample the $\{\gamma_{1:C}\}$ from their full conditional distribution. This ensures that the reduced conditioning performed in the M-H step does not change the stationary distribution of the Markov chain \citep{VanDyk2015} \--- see Appendix \ref{app:GibbsSamp}.

\end{enumerate}

\subsection{Sequential Posterior Approximation}
\label{seqpriors}

In flow cytometry experiments it is common to actually have multiple datasets $\boldsymbol{y}^{(i)}$ (with $i=1,\dots, I$)  corresponding to multiple individuals, or repeated measurements of the same individual. In such cases, it is of interest to use previous time points or previous samples results as prior information, in order to leverage all the information available to estimate the mixture. However, specifying prior information to Dirichlet process mixture models is not straightforward \citep{Kessler2015}. Here we propose to use the posterior MCMC draws obtained from previous dataset $\boldsymbol{y}^{(i)}$ as prior information to analyze the next dataset $\boldsymbol{y}^{(i+1)}$. To do so, first let's consider the hierarchical model using all observations from both $\boldsymbol{y}^{(i)}$ and $\boldsymbol{y}^{(i+1)}$ at once :

\begin{subequations}
\begin{align}
\alpha & \sim Gamma(a,b)\\
G | \alpha & \sim DP(\alpha,G_{0})\\
\boldsymbol{y}^{(i)},\boldsymbol{y}^{(i+1)}|G  & \overset{\tiny{\iid}}{\sim}\int_{\boldsymbol{\Theta}} f_{\boldsymbol{\theta}}(\cdot)dG(\boldsymbol{\theta})
\end{align}
\label{eq:model2datasets}
\end{subequations}

\noindent We are interested in estimating $p(G|\boldsymbol{y}^{(i)},\boldsymbol{y}^{(i+1)})\propto p(G|\boldsymbol{y}^{(i)})p(\boldsymbol{y}^{(i+1)}|G)$. The idea is to first approximate $p(G|\boldsymbol{y}^{(i)})$ by a Dirichlet process through MCMC draws from the model described in \ref{ss:model}:

\begin{equation}
p(G|\boldsymbol{y}^{(i)})\simeq\int DP(G;\alpha,G_{1})Gamma(\alpha;a_{1}%
,b_{1})d\alpha
\label{eq:modelSeqPost}
\end{equation}

\noindent where $G_{1}$, $a_{1}$, $b_{1}$ are parameters to be estimated from the MCMC approximation of the true posterior: i) $\widehat{a_{1}}$ and $\widehat{b_{1}}$ can be taken as MLE estimates from the MCMC samples $\alpha^{(j)}$ ; ii) $\widehat{G_{1}}$ is a parametric approximation of the posterior mixing distribution $G_1$ (the true posterior is not suitable for being directly plugged in as a base distribution parameter of another $DP$ as it is nonparametric). In the case of a skew t-distributions mixture model, we approximate $G_1$ with the following joint distribution: $G_1\simeq(sNiW,P_{0,\nu})$ where $P_{0,\nu}$ is the chosen prior for the skew t-distribution degrees of freedom. To estimate $G_1$, we estimate the Maximum  \textit{a posteriori} (MAP) from the posterior MCMC samples (see Appendix \ref{app:ParamEst}).

Now using this posterior parametric approximation, we have the same hierarchical model as before but conditional on $\boldsymbol{y}^{(i)}$:

\begin{subequations}
\begin{align}
\alpha | \boldsymbol{y}^{(i)}  & \sim Gamma(\widehat{a_{1}},\widehat{b_{1}})\\
G|\alpha, \boldsymbol{y}^{(i)}  & \sim DP(\alpha,\widehat{G_{1}})\\
\boldsymbol{y}^{(i+1)} | G, \boldsymbol{y}^{(i)}  & \overset{\tiny{\iid}}{\sim}\int_{\boldsymbol{\Theta}} f_{\boldsymbol{\theta}}(\cdot)dG(\boldsymbol{\theta})
\end{align}
\label{eq:model2ndPost}
\end{subequations}

Note that under this approximate posterior model, the cluster parameters $\boldsymbol{\theta}_{k}^*$ are \iid~from $G_{1}$. Such an approach can be iterated a number of times, if for instance several time points are observed, iteratively approximating the successive posteriors. This approach allows to finally account for all the previous information in the mixture model estimation.

\subsection{Point estimate of the clustering}

Getting a representation of the partition posterior distribution is difficult \citep{Medvedovic2002}. One can use the maximum a posteriori, i.e. using the point estimation form the MCMC sample that maximize the posterior density. However this ignores all the information about the uncertainty around the partition gained through the Bayesian approach.

Another way is to rather consider a co-clustering posterior probability (or similarity) matrix $\zeta$ on each pair $(c,d)$ of observations. Such a matrix can be estimated by averaging the co-clustering matrices from all the explored partitions in the posterior MCMC draws:
\begin{equation}
	\widehat{\zeta}_{cd}=\frac{1}{N}\sum_{i=1}^N\delta_{\ell_c^{(i)}\ell_d^{(i)}}
\end{equation}
where $N$ is the number of MCMC draws from the posterior and $\delta_{kl}=1$ if $k=l$, $0$ otherwise.
An optimal partition point estimate $\{\widehat{\ell}_{1:C}\}$ can then be derived in regard of this similarity matrix through stochastic search with the explored partitions in the posterior MCMC draws \citep{Dahl2006}, by using a pairwise coincidence loss function \citep{Lau2007} such as the one proposed by \cite{Binder1978, Binder1981} which optimizes the Rand index \citep{Fritsch2009a}:
\begin{equation}
\{\widehat{\ell}_{1:C}\}  = \underset{\{\ell^{(i)}_{1:C}\}\in\left\{\{\ell^{(1)}_{1:C}\}, \dots, \{\ell^{(N)}_{1:C}\}\right\}}{\operatorname{arg\,min}} \sum_{c=1}^{C-1}\sum_{d=c+1}^C2\left(\delta_{\ell_c^{(i)}\ell_d^{(i)}} - \widehat{\zeta}_{cd}\right)^2
\end{equation}
The computational cost of this approach, though, is of the order $\mathcal{O}(NC^2)$ due to the necessity of computing all the similarity matrices.

A different optimal partition point estimate $\{\tilde{\ell}_{1:C}\}$ can also be derived using the $\mathcal{F}$-measure as our loss function. The $\mathcal{F}$-measure is widely used as a way to summarize the accordance between 2 methods, one being considered as a reference (gold-standard). It is the harmonic mean of the precision and recall:
\begin{equation}
	\mathcal{F}=\dfrac{2 Pr Re}{Pr+ Re}
\end{equation}
In order to use the $\mathcal{F}$-measure to evaluate our clustering method, we rely on the definition proposed in the online methods from \cite{Aghaeepour2013}. In this setting of unsupervised clustering, the precision $Pr$ is the number of cells correctly assigned to a given cluster divided by the total number of cells assigned to that cluster (also called Positive Predictive Value). The recall $Re$ is the number of cells correctly assigned to a given cluster divided by the number of cells that should be assigned to this cluster according to the gold-standard. Since in our problem the labels of the different clusters are exchangeable, the $\mathcal{F}$-measure is computed for each combination of the reference clusters and the predicted clusters. Let $G=\{g_1,\dots, g_m\}$ be a set of $m$ reference clusters and $H=\{h_1,\dots,h_n\}$ be set of $n$ predicted clusters. For each combination pair $(q,r)$ of a reference cluster $g_q$ and a predicted cluster $h_r$, the $\mathcal{F}$-measure is computed as follows:
\begin{equation}
 Pr(h_r, g_q)= \dfrac{|g_q\cap h_r|}{|h_r|}\quad \text{and}\quad Pr(h_r, g_q)= \dfrac{|g_q\cap h_r|}{|g_q|}
\end{equation}
\begin{equation}
 \mathcal{F}(h_r, g_q)=\dfrac{2 Pr(g_q, h_r) Re(g_q, h_r)}{Pr(g_q, h_r)+ Re(g_q, h_r)}
\end{equation}
This $\mathcal{F}$-measure is comprised in $[0,1]$, and the closer it is to 1 the better the agreement is between the predicted cluster and the reference cluster. The total $\mathcal{F}$-measure for a predicted partition $H$ given a gold-standard $G$ is then define as the weighted  sum of the best matched $\mathcal{F}$-measure: 
\begin{equation}
	\mathcal{F}_{tot}(H, G) = \frac{1}{\sum_{q=1}^m |g_q|}\sum_{q=1}^m |g_q|\,\underset{r\in\{1,\dots,n\}}{\operatorname{max}}\,\mathcal{F}(h_r, g_q)\label{globalF}
\end{equation}
This total $\mathcal{F}$-measure is again between $0$ and $1$, and the closer it is to 1 the better the predicted partition agrees with the gold-standard. The optimal partition point estimate in respects of this $\mathcal{F}$-measure is then obtained with the partition that maximizes its average $\mathcal{F}$-measure over all the other explored partitions in the posterior MCMC draws:
\begin{equation}
	\{\tilde{\ell}_{1:C}\} = \underset{\{\ell^{(i)}_{1:C}\}\in\left\{\{\ell^{(1)}_{1:C}\}, \dots, \{\ell^{(N)}_{1:C}\}\right\}}{\operatorname{arg\,max}} \,\frac{1}{N} \sum_{\substack{j=1 \\ j\neq i}}^N \mathcal{F}_{tot}\left(\{\ell^{(i)}_{1:C}\}, \{\ell^{(j)}_{1:C}\}\right)
\end{equation}
Note the $\mathcal{F}$-measure is computed here only between sampled partitions, and a gold-standard partition is unnecessary.

%\begin{equation}
%	p(\nu_k)\propto \left[\left(\frac{\nu}{\nu+3}\right)\left(\varphi'\left(\frac{\nu}{2}\right) - \varphi'\left(\frac{\nu+1}{2}\right) - \frac{2(\nu+3)}{\nu(\nu+1)^2}\right)\right]^{\frac{1}{2}}
%\end{equation}
%where $\varphi'$ is the trigamma function.

\section{Simulations Study}

\subsection{Non informative prior}
\begin{figure}[!h]
	\centerline{\includegraphics[width=0.85\textwidth]{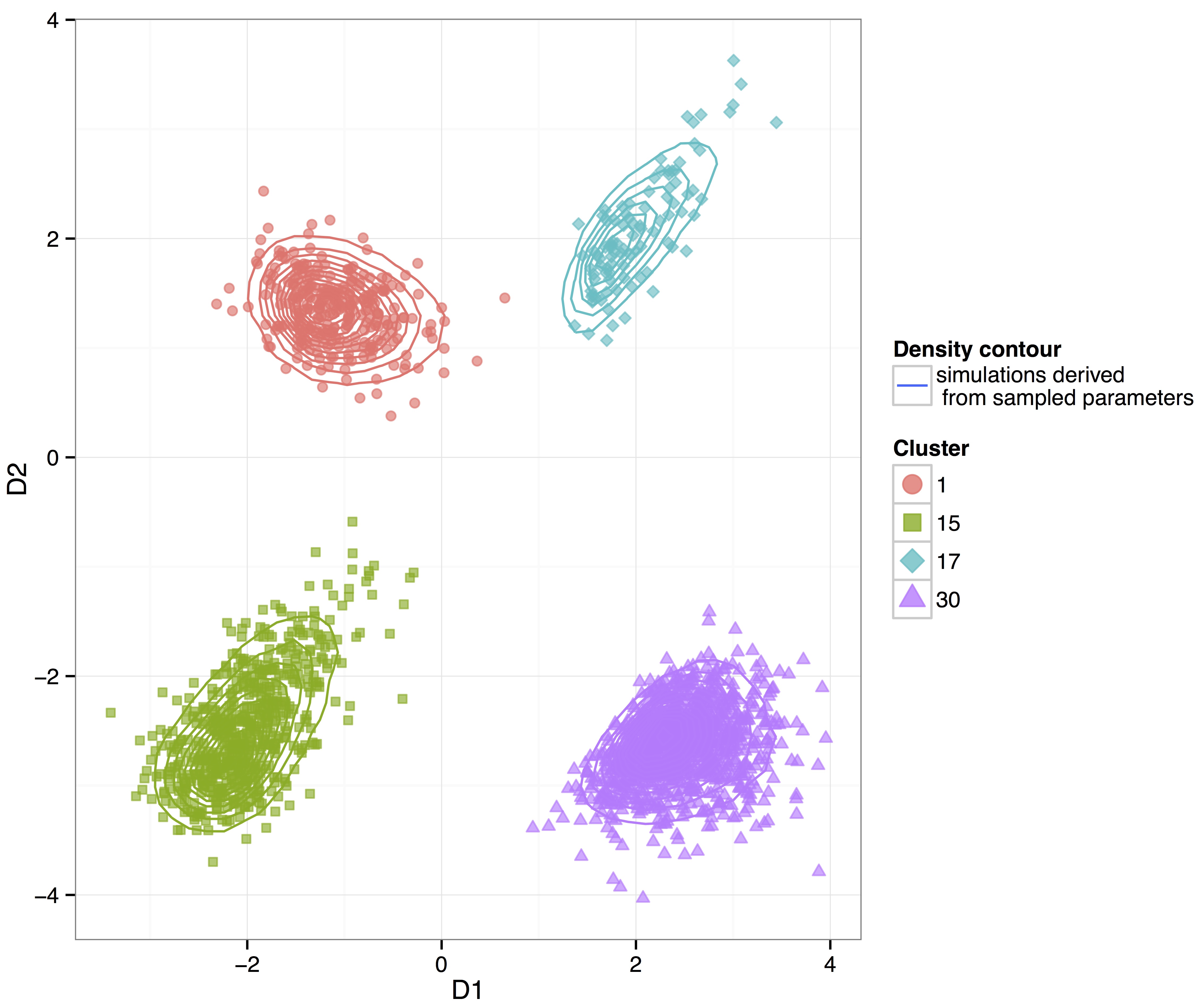}}
	\caption{Partition point estimate from one of the first 100 2-dimensional simulations}
	\label{fig:FitSimuEx}
\end{figure}

First, to assess the performances of the Dirichlet process mixture of skew $t$ distributions model in a simple clustering case, 100 simulations in 2-dimensions were performed. In each simulation 2000 observations were drawn from 4 distinct clusters representing respectively 50\%, 30\%, 15\% and 5\% of the data. After 10,000 MCMC iterations (9,000 iterations burnt and a thining of 5 gave 200 partitions sampled from the posterior; the chain was initialized with 30 clusters), the resulting mean $\mathcal{F}$-measure was  0.998 when comparing the partition point estimate obtained from our approach with the true clustering of the simulated data. In 97\% of the cases, the partition point estimate had 4 clusters (i.e. the true number of clusters in the simulated data), while it had 3 or 5 clusters in the remaining 3\%. Figure \ref{fig:FitSimuEx} shows an example of the partition point estimate obtained for one of those simulation run.

\subsection{Sequential posterior approximation plugged-in as informative prior}
\label{ss:seqpostsims}

To illustrate how the sequential posterior approximation strategy compares to the standard non informative prior setting, we ran simulations where we considered two samples derived from the same infinite mixture model. The first sample is simulated for a time $t$, and the second sample at $t+1$. As all observations originate from the exact same distribution, regardless of the sample, the hypothesis of the sequential posterior approximation strategy is satisfied. One of the major gain observed is the time to convergence for the partition%, as exhibited in Table \ref{tab:conv}
. Using an informative prior derived from the sample at time $t$ to estimate the partition of the sample from $t+1$ makes it more than three time faster to converge according to the Gelman-Rubin statistics%for the Gelman-Rubin statistic to drop below the 1.05 threshold
.

%\begin{table}
%\begin{center}
%\caption{Average number of MCMC iterations needed before convergence$^\dagger$ \label{tab:conv} of the concentration parameter $\alpha$}
%\begin{tabular}{lr}
%\hline
%Standard & 1,743 \\
%Informative prior & 502 \\
%\hline\smallskip\\
%\end{tabular}
%\end{center}
%\footnotesize$\dagger$: Gelman-Rubin statistics drops below 1.05
%\end{table}

\begin{figure}[!h]
	\centerline{\includegraphics[width=0.85\textwidth]{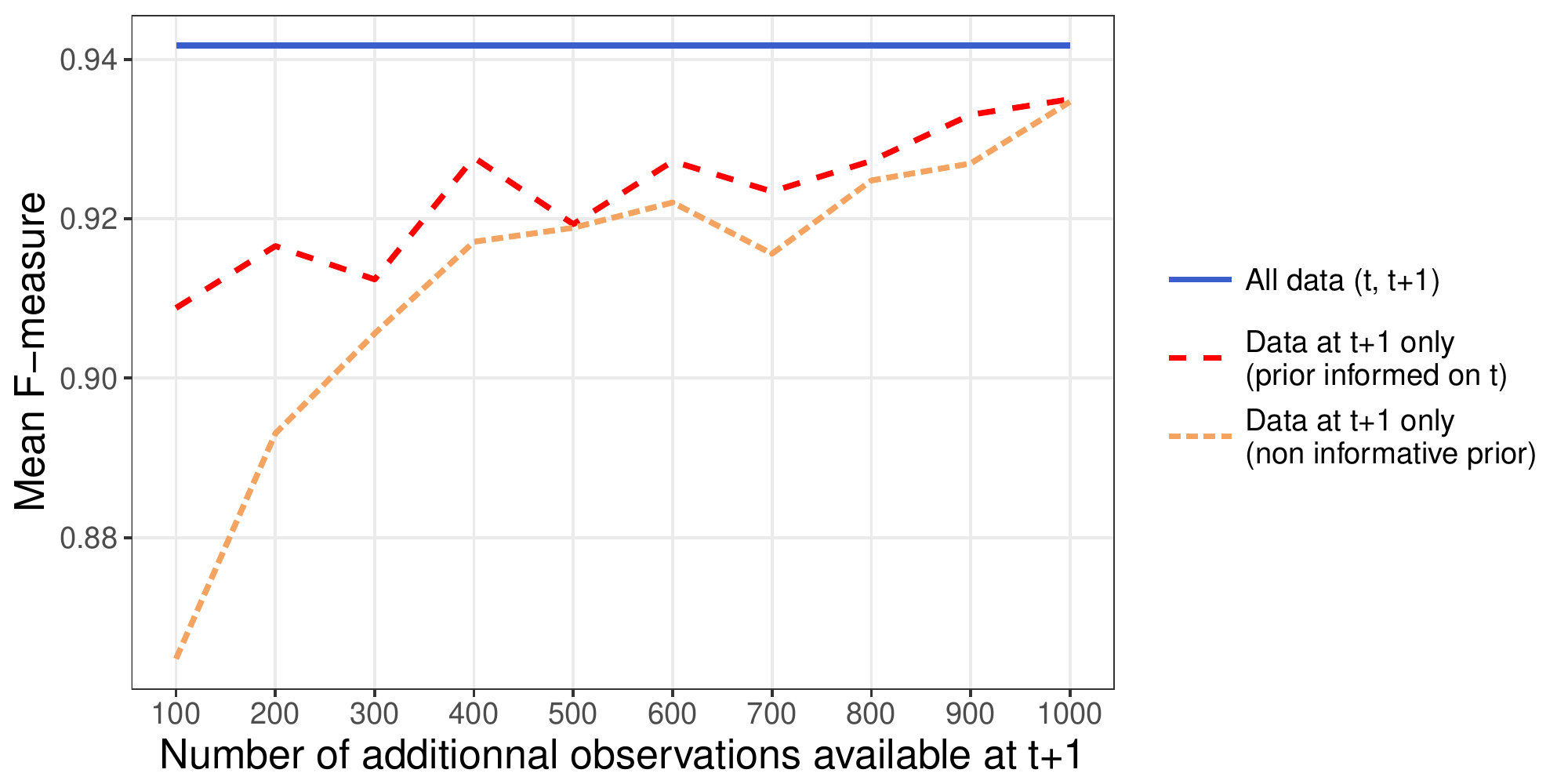}}
	\caption{Mean $\mathcal{F}$-measure according to the number of observations available at $t+1$, while 1,000 observations are available at $t$, over 300 simulations}
	\label{fig:FseqMean}
\end{figure}

In further simulations, we also investigated the performance of this sequential posterior approximation strategy.  As opposed to using the standard non informative prior, it shows substantial gains when the amount of information brought by the prior is substantial compared to the amount available from the data at $t+1$ alone. As the amount of information available at $t+1$ increases, the gain from using this strategy can become less noticeable, as shown using the $\mathcal{F}$-measure in Figure \ref{fig:FseqMean}. But even when as many observations are available at $t+1$ as at $t$, the accuracy for rare cell populations is still improved by using an informative prior. This is not necessary visible at the scale of the total $\mathcal{F}$-measure, because it is masked by the larger clusters. However, when computing a limited $\mathcal{F}$-measure, that only takes into account smaller clusters (see Appendix \ref{app:limitedF}), the use of an informative prior in this sequential strategy seems to always improves the clustering accuracy for smaller clusters (see Supplementary Figure \ref{fig:FLimited} in Appendix \ref{app:limitedF}).

%\begin{figure*}
%\centerline{\includegraphics[width = 0.85\linewidth]{RawDataSeqPriorEx}\vspace{0.6cm}}
%\centerline{\includegraphics[width = 0.85\linewidth]{ResBroadAndSeqprior}\vspace{0.6cm}}
%\centerline{\includegraphics[width = 0.85\linewidth]{HMBroadAndSeqprior.pdf}}
%\caption{Example from one of the second set 100 2-dimensional simulations}
%\label{fig:SeqPriorSimuEx}
%\end{figure*} Because the sequential posterior approximation strategy also improves the parameter space exploration by the MCMC chain, it exhibits a substantial speed up in reaching the invariant probability distribution, as displayed in Figure \ref{fig:convLogPost}.

%\begin{figure*}
%	\centerline{\includegraphics[width=0.9\textwidth]{ConvLogPost}}
%	\caption{Trace of the log-posterior exhibiting accelerated convergence when using sequential posterior approximation}
%	\label{fig:convLogPost}
%\end{figure*}

\section{Application to real datasets}
\subsection{Benchmark Graft versus Host Disease dataset}

The Graft versus Host Disease (GvHD) dataset is a public dataset that was first analysed (manually gated) in \cite{Brinkman2007}, with the objective of identifying cellular signature that correlates or predict Graft versus Host disease. The GvHD data were used as benchmark data in the FlowCAP challenge \cite{Aghaeepour2013}. Flow cytometry data was collected for 12 sample, and original manual gates are being regarded as the true cell clustering (actually a consensus over eight manual operators, from eight different operators). In order to try to mitigate further the well known reproducibility issues with manual gating \citep{Ge2012, Aghaeepour2013}, only the most concordant clusters between the 8 gatings ($\mathcal{F}$-measure above 0.8) were used for comparing with the automated results, as was done in \cite{Aghaeepour2013}. 
The data were downloaded from the FlowCAP project website [\footnotesize{\url{http://flowcap.flowsite.org/}}\normalsize] as part of the FlowCAP-I challenge [\footnotesize{\url{http://flowcap.flowsite.org/codeanddata/FlowCAP-I.zip}}\normalsize]. Table \ref{tab:FmeasAghaeepour} shows the performance of our proposed approach NPflow on this dataset, in the context of the other approaches reviewed by \cite{Aghaeepour2013}. The $\mathcal{F}$-measure is computed for all samples available for a given dataset and the mean over all samples is reported, as well as bootstrap 95\% Confidence Intervals. No algorithm is performing significantly better than NPflow thus placing NPflow among the top methods for automatic gating, and the sequential approach yields a $\mathcal{F}$-measure higher than any other method.

\begin{table}
\begin{center}
\caption{Mean $\mathcal{F}$-measures across all the 12 samples from the GvHD benchmark dataset\label{tab:FmeasAghaeepour}}
%\vspace*{0.5cm}
\begin{tabular}{lc}
\hline
Method & $\mathcal{F}$-measure\\
\hline
NPflow & 0.85 (0.80, 0.90)\\
NPflow-seq & 0.89 (0.85, 0.94)\\
ADICyt & 0.81 (0.72, 0.88)\\
CDP & 0.52 (0.46, 0.58)\\
FLAME & 0.85 (0.77, 0.91)\\
FLOCK & 0.84 (0.76, 0.90)\\
flowClust/Merge & 0.69 (0.55, 0.79)\\
flowMeans & 0.88 (0.82, 0.93)\\
FlowVB & 0.85 (0.79, 0.91)\\
L2kmeans & 0.64 (0.57, 0.72)\\
MM & 0.83 (0.74, 0.91)\\
MMPCA & 0.84 (0.74, 0.93)\\
SamSPECTRAL & 0.87 (0.81, 0.93)\\
SWIFT & 0.63 (0.56, 0.70)\\
\hline
\end{tabular}
\end{center}\bigskip\smallskip
\footnotesize All estimates except for our proposed NPflow approach are from \cite{Aghaeepour2013}. 95\% Confidence Intervals are calculated on 10,000 bootstrap samples of the $\mathcal{F}$-measures.
\end{table}

The GvHD benchmark data are not longitudinal data. However, the sequential posterior model can still improve the results by using each individual sample sequentially. Using this dataset, the mean $\mathcal{F}$-measure reached 0.89 (0.85, 0.94) with the sequential approach, compared to a value of 0.85 (0.80, 0.90) with the standard NPflow model (Table \ref{tab:FmeasAghaeepour}). The sequential strategy exhibits the highest $\mathcal{F}$-measure for the GvHD dataset, making it the best approach for unsupervised automatic gating compared to competing methods evaluated in \cite{Aghaeepour2013}.

\subsection{Original DALIA-1 data}
\label{ss:dalia}

\begin{figure}[!h]
	\centerline{\includegraphics[width=0.8\textwidth]{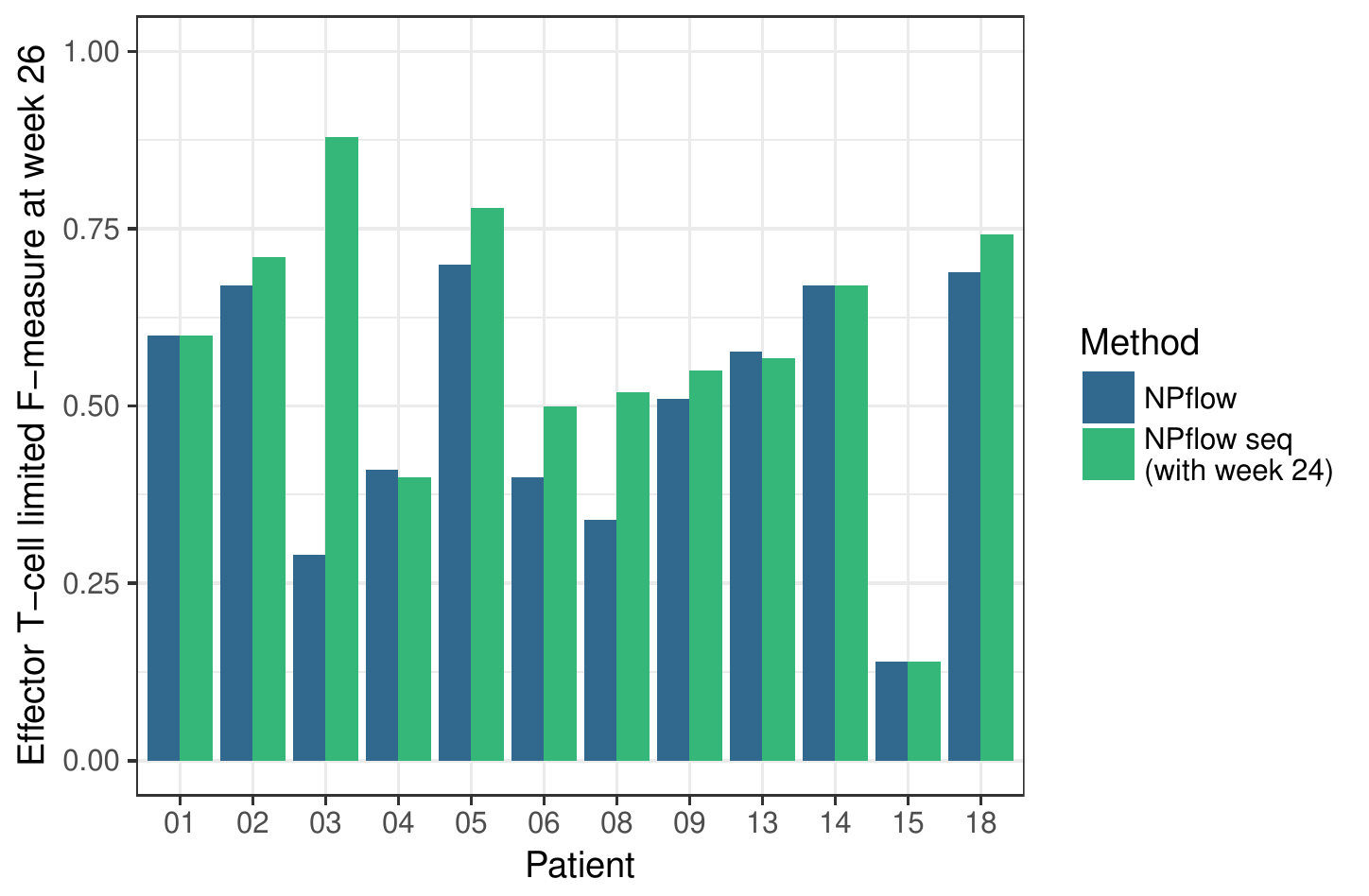}}
	\caption{Limited $\mathcal{F}$-measures for the effector CD4+ T-cell population from the DALIA-1 trial two weeks after HAART interruption for NPflow with or without the sequential strategy, compared to manual gating.}
	\label{fig:FmeasDalia}
\end{figure}

\begin{figure}[!h]
	\centerline{\includegraphics[width=0.95\textwidth]{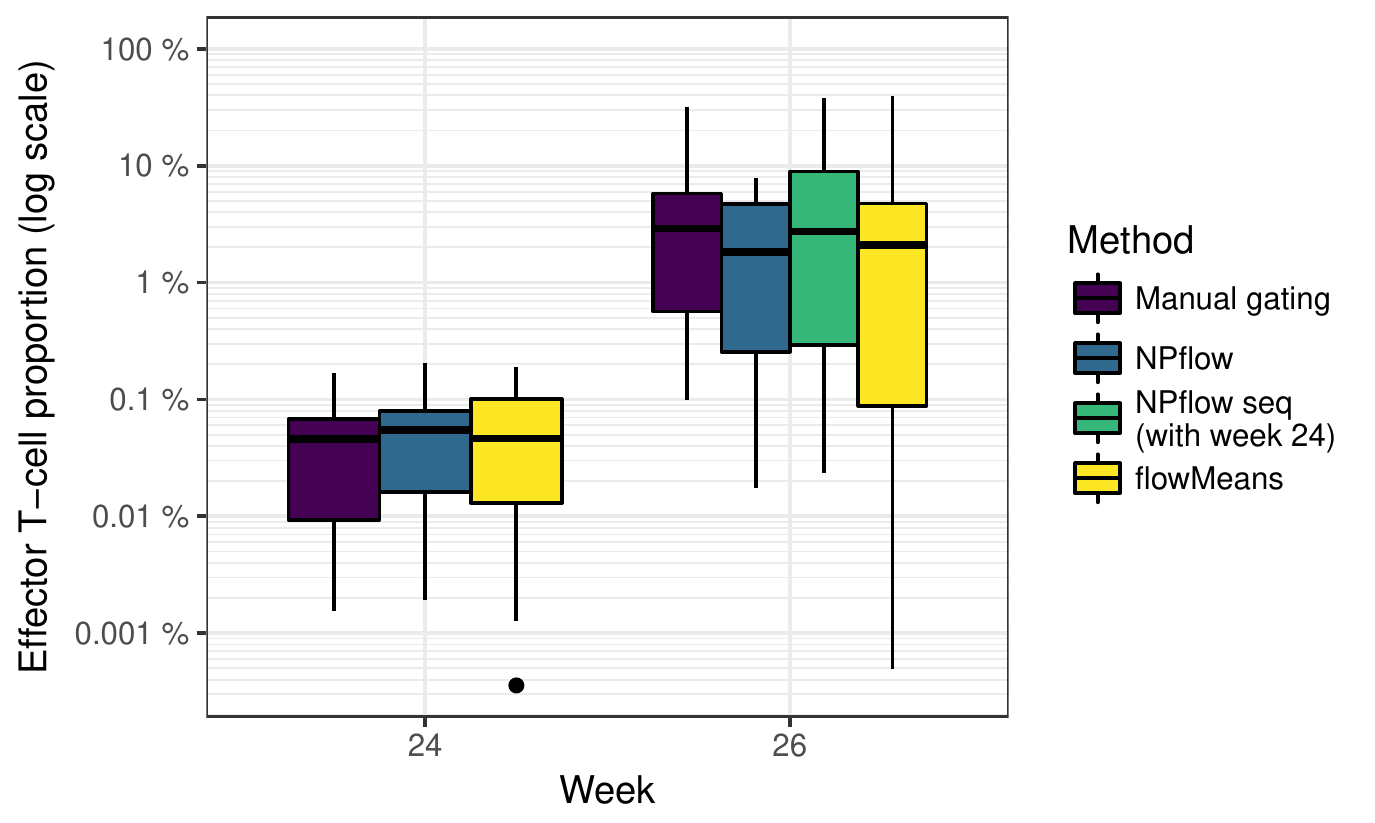}}
	\caption{Evolution of the proportion of effector CD4+ T-cells in the DALIA-1 trial following HAART interruption (from manual gating).}
	\label{fig:PropDalia_DescAll}
\end{figure}

We also applied our method to an original dataset from DALIA-1, a phase I trial evaluating a therapeutic vaccine against HIV \citep{Levy2014}. The vaccine candidate was based on ex-vivo generated interferon-$\alpha$ dendritic cells loaded with HIV-1 lipo-peptides, and activated with lipopolysaccharide. The objectives of the trial were to evaluate the safety of the strategy and to evaluate the immune response to the vaccine. For our purpose here, we are interested in the 12 HIV positive patients who had their cellular populations quantified at 18 time-points during the trial. More specifically, we focused on two time points (at week 24 and week 26 of the trial) immediately following antiretroviral treatment (HAART) interruption which took place at week 24. Following this interruption, the increase of viral replication is associated with changes in cell populations \citep{Thiebaut2005, Levy2012b}. Here we especially looked at the effector CD4+ T-cells, defined as CD45RA+CD27- among the CD3+CD4+ cells \citep{Larbi2014}, that are one of the first cell populations to be affected during the viral rebound \citep{Levy2012b}. Since flow-cytometry measurements were repeated at each time points for each patients, we used the sequential strategy at week 26, in the hope to use the information from week 24 to better identify the effector CD4+ T-cell population at the next time point. Figure \ref{fig:FmeasDalia} illustrates the overall efficiency gain at week 26 from using the sequential strategy. The average limited $\mathcal{F}$-measure (compared to available manual gating used as gold-standard) on those 12 samples is $0.5$ for NPflow with a non-informative prior, and increase to $0.59$ with the sequential strategy. By comparison, flowMeans (the second best method on the GvHD dataset) gives an average limited $\mathcal{F}$-measure of $0.51$ (see Appendix \ref{app:FM} for details).  Figure \ref{fig:Pat3_ex} gives an example of a patient for which the sequential strategy was especially improving the identification of the effector CD4+ T-cells. In this case, the percentage of effector CD4+ T-cells was estimated at 31.7 by the manual gating, at 7.5 by NPflow, and at 38.1 by the sequential strategy. Figure \ref{fig:PropDalia_DescAll} shows the general increase of effector CD4+ T-cell proportions for every patients after treatment interruption (see Appendix \ref{app:FM} for more details).

\begin{figure}[!h]
	\centerline{\includegraphics[width=0.85\textwidth]{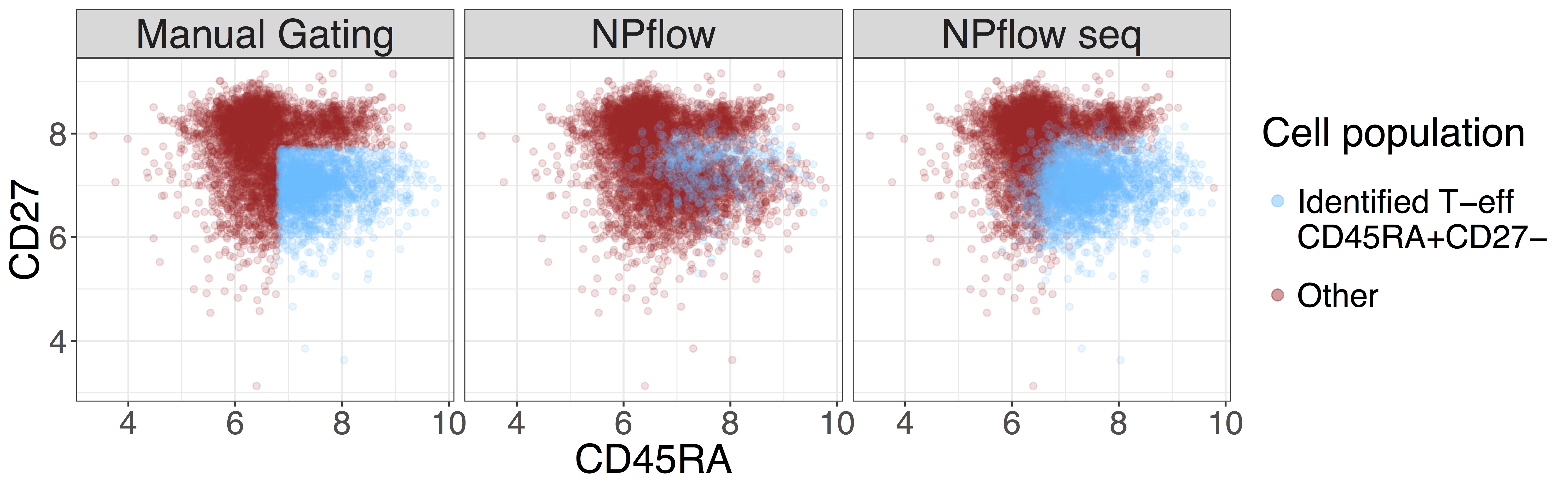}}
	\caption{ CD3+CD4+ cells of patient 3 from the DALIA-1 trial two weeks after HAART interruption (at week 26).}
	\label{fig:Pat3_ex}
\end{figure}

\begin{figure}[!h]
\centerline{\includegraphics[width=0.85\textwidth]{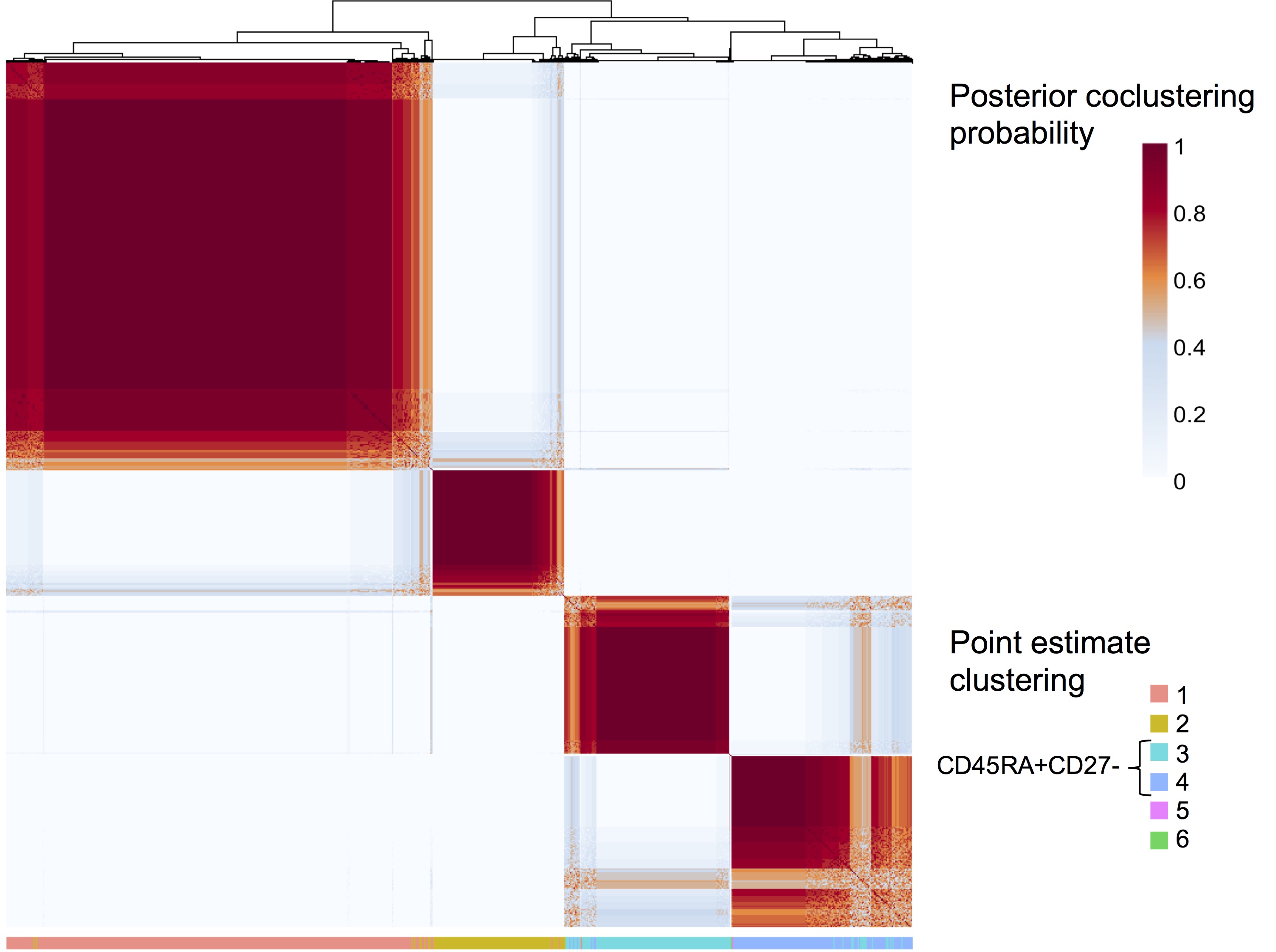}}
\caption{Heatmap of the posterior co-clustering probabilities for the CD3+CD4+ cells of patient 3 at week 26 from DALIA-1.}\label{simMat}
\end{figure}

In addition to providing a point estimate of the partition, our method also quantifies the uncertainty around the posterior clustering through posterior co-clustering probabilities. Figure \ref{simMat} displays such a co-clustering posterior probabilities matrix. Where we can clearly identify 4 core clusters, with some uncertainty corresponding to marginal cells that are in between overlapping populations.

\section{Discussion}
\label{s:discuss}

We extend the classical Dirichlet process Gaussian mixture model to skew t-distribution mixtures, based on \cite{Fruhwirth-Schnatter2010a} parametrization of such distributions. Such an approach is well suited for unsupervised model based classification of flow cytometry data. Automatic gating of cell populations is an open research problem and the proposed approach features two important characteristics for this task: i) it avoids the difficult issue of model selection by estimating directly the number of components in the mixture ; ii) it uses skew and heavy tailed distributions in the form of skew t-distributions, of which the gaussian is a particular case. Estimation of the posterior co-clustering probabilities for each data pair allows to quantify the uncertainty about the posterior partition, and an optimal point estimate of the clustering is provided by minimizing a cost function in regards to the average posterior co-clustering matrix. We have developed and implemented an efficient collapsed Metropolis within Gibbs sampler for estimating such models. One of the advantage of our proposed sampler is the absence of label switching issue, as it uses directly the partition of the data without having to deal with labels \citep{Jasra2005}. As an indication of runtime, around 3,000 MCMC iterations can be run on average for a real dataset of around 10,000 observations over 6 dimensions, using one Intel\textregistered~Xeon\textregistered~x5675 processor in an hour. Besides, instead of using a partially collapse Gibbs sampler algorithm, it could be of interest to also investigate the use of sequential Monte-Carlo algorithms, especially for the sequential modeling strategy or other possible dynamic extensions of the model proposed here \citep{Caron2008, Caron2017}.

We propose to use sequential parametric approximations of the posterior as refined informative priors in case of repeated measurement of flow cytometry data. The proposed sequential analysis strategy enables to analyze each sample sequentially, as the data are acquired. It does not require to wait for the last sample to perform the automatic gating nor to analyze all data at once, but it still uses available prior knowledge. This contrasts with hierarchical extensions of the Dirichlet Process Mixture Model such as those proposed by \cite{Cron2013} or \cite{Dundar2014}, where the complete dataset must be analyzed at once. This sequential strategy allows one to analyze the samples as they are acquired, which can be useful in clinical trials where there are often intermediate analyses for instance. Moreover in large studies the size of the data can make it challenging to analyze all samples at once, and such a sequential approach then makes practical sense \citep{Huang2005a}. Futhermore, this use of sequentially informed priors does not face the usual complications of cluster matching arising when an algorithm is run on each sample separately \citep{Cron2013}. In our simulation study this sequential posterior approximation strategy improves the fit of the model. In addition, such a strategy exhibits accelerated convergence and greater accuracy for small clusters, as long as the different samples are similar enough. Besides, the parametric prior can also be specified to inform the model with expert knowledge, e.g. to favor a range for the expected number of clusters. On real flow-cytometry data we show that the sequential strategy also improves the clustering performances. On the benchmark GvHD dataset, it outperforms all other methods investigated in by \cite{Aghaeepour2013}. In the DALIA-1 trial, the sequential strategy allows a better recovery of the effector CD4+ T-cell population after a important perturbation of this targeted population following HAART interruption among HIV positive patients. It is worth noting however that in other cases, for instance if the data distributions were too different from samples to samples, the sequential posterior model would not necessarily improve the clustering results, and could even gave a diminished $F$-measure compared to the non sequential strategy.

Manual gating is still considered the gold-standard when evaluating an automatic gating strategy on real flow cytometry data. Yet one should keep in mind that manual gating has reproducibility issues, often resulting in a partial and subjective clustering \citep{Ge2012, Welters2012, Aghaeepour2013, Gondois-Rey2016}. Therefore using manual gating as the gold-standard might not be actually the best way to assess the performance of automatic gating algorithms on real data, because of its inherent flaws.

Mass cytometry is a technology very similar to flow cytometry. Using ions in place of colors, CyTOF is able to measure up to 40 cell markers at once, generating even more data than flow cytometry. Efficient automated gating method are therefore all the more needed in the context of CyTOF\citep{Melchiotti2016}. The approach proposed here could be directly applied to such data. More generally, we propose here a framework for Dirichlet process mixtures of multivariate skew t-distributions modeling that is suitable for any kind of data modeled as such a mixture, especially when the number of mixture component is unknown. We provide an efficient implementation of our method within the R package \texttt{NPflow} that is available on CRAN at \url{https://cran.r-project.org/web/packages/NPflow}.

\section*{Software}
\label{soft}

Software in the form of R code is available on the Comprehensive R Archive Network as an R package \texttt{tcgsaseq}.

\section*{Acknowledgements}

The authors are extremely grateful to Jean-Louis Palgen for his time and efforts to manually gate the effector CD4+ T-cells in the DALIA-1 trial at weeks 24 and 26, as well as to the DALIA-1 study group. The authors also thank Nima Aghaeepour for his help in using supplementary data provided with his publication \citep{Aghaeepour2013}. Part of this work has been supported by the BNPSI ANR project no ANR-13-BS-03-0006-01. Boris P. Hejblum was a recipient of a Ph.D. fellowship from the \'Ecole des Hautes \'Etudes en Sant\'e Publique (EHESP) Doctoral Network. Part of this work has been supported by the IMI2 grant EBOVAC2. Computer time for this study was partly provided by the computing facilities MCIA (M\'esocentre de Calcul Intensif Aquitain) of the Universit\'e de Bordeaux and of the Universit\'e de Pau et des Pays de l'Adour.\vspace*{-8pt}

\bigskip

\noindent {\it Conflict of Interest}: None declared.

\bibliographystyle{biom}
\bibliography{cyto}

\section*{Appendix}
\label{supMat}
\setcounter{section}{0}
\renewcommand{\thesection}{\Alph{section}}
\setcounter{figure}{0}
\renewcommand{\thefigure}{S\arabic{figure}}

\section{Gibbs samplers}
\label{app:GibbsSamp}

\begin{itemize}
\item $K$ is the number of different unique values taken by $c$ (i.e. the number of clusters). This number of clusters $K$ is not set and its value may change at each iteration. 
\item $\ell_{c}$ is the latent variable indicating which cluster the observation $c$ belongs to. $\{\ell_{1:C}\}$ refers to a whole partition of the data. 
\item $s_c$ is the skew parameter for the observation $c$.
\item $\gamma_c$ is the scale parameter (skew t only) for the observation $c$.
\end{itemize}\medskip

\subsection{Skew Normal distributions mixture}
Our Gibbs sampler proceeds with each of the following updates in turn:

\begin{enumerate}
	
	\item update concentration parameter $\alpha$ given $\{\ell_{1:C}\}$ using the data augmentation technique from \cite{West1992}:\\
		$\alpha \propto p(\alpha | \{\boldsymbol{z}_{1:C}\}, G_0, \{\boldsymbol{\xi}_k\}, \{\boldsymbol{\psi}_k\}, \{\boldsymbol{\Sigma}_k\}, \{\ell_{1:C}\}, \{w_k\}, \{s_{1:C}\}) \propto p(\alpha |\{\ell_{1:C}\})$\\
		$(\alpha,x | \{\ell_{1:C}\})\sim p(\alpha)\alpha^{K-1}(\alpha+C)x^\alpha(1-x)^{C-1}$\\
		$(x | \alpha, \{\ell_{1:C}\}) \sim \text{Beta}(\alpha+1, C)$\\
		$(\alpha | x, \{\ell_{1:C}\}) \sim \pi_x \text{Gamma}(a + K, b-\log(x)) + (1-\pi_x) \text{Gamma}(a + K-1, b-\log(x))$ with $p(\alpha)\propto \text{Gamma}(a,b)$ and $\frac{\pi_x}{1-\pi_x}=\frac{a+k-1}{C(b-\log(x))}$

	\item update $G$ given $\alpha$,  $\{\boldsymbol{\xi}_k\}$, $\{\boldsymbol{\psi}_k\}$, $\{\boldsymbol{\Sigma}_k\}$ and $G_0$ via slice sampling:\\
	$\{w_k\} , \{\ell_{1:C}\} \propto p\left(\{w_k\}, \{\ell_{1:C}\}| \{\boldsymbol{z}_{1:C}\}, \alpha, G_0, \{\boldsymbol{\xi}_k\}, \{\boldsymbol{\psi}_k\}, \{\boldsymbol{\Sigma}_k\}, \{s_{1:C}\}\right)$
	
	\begin{enumerate}
		\item sample the weights:\\
		$(w_1, \dots, w_K, w_* | \{\ell_{1:C}\}) \sim \text{Dirichlet} (\text{card}(\{\ell_c=1\}), \dots, \text{card}(\{\ell_c=K\}), \alpha)$
	
		\item for $c=1,\dots, C$: 
		$u_c\sim\text{Unif}([0, w_{\ell_c}[)$
		
		\item Set $j=K$. While $\sum_{k=1}^j w_k < (1-min(u_{1:C}))$:
			\begin{itemize}
				\item set $j=j+1$
				\item sample $\pi_j\sim\text{Beta}(1,\alpha)$
				\item set $w_j =w_*\pi_j\prod_{k={K+1}}^{j-1}(1-\pi_k)$
				\item sample $(\boldsymbol{\xi}_j, \boldsymbol{\psi}_j, \boldsymbol{\Sigma}_j | G_0)\sim G_0$
			\end{itemize}
		
		\item for $c=1,\dots,C$ sample $\ell_c$ given $\{\boldsymbol{\xi}_k\}$, $\{\boldsymbol{\psi}_k\}$, $\{\boldsymbol{\Sigma}_k\}$, $\{w_k\}$ from:\\$p(\ell_c=k) \propto \mathds{1}_{\{w_{k} > u_c\}}f_{\mathcal{SN}}(\boldsymbol{z}_c, \boldsymbol{\xi}_k, \boldsymbol{\psi}_k, \boldsymbol{\Sigma}_k)$
	\end{enumerate}

		\item for $c=1,\dots,C$  update $s_c$ given $\ell_c$ , $\{\boldsymbol{\xi}_k\}$, $\{\boldsymbol{\psi}_k\}$, $\{\boldsymbol{\Sigma}_k\}$:\\
		$p(s_c | \boldsymbol{z}_c, \alpha, G_0, \{\boldsymbol{\xi}_k\}, \{\boldsymbol{\psi}_k\}, \{\boldsymbol{\Sigma}_k\}, \{\ell_{1:C}\}, \{w_k\})$ $\propto p(s_c | \boldsymbol{z}_c , \{\boldsymbol{\xi}_k\}, \{\boldsymbol{\psi}_k\}, \{\boldsymbol{\Sigma}_k\}, \ell_c)$\\ $(s_c | \boldsymbol{z}_c , \{\boldsymbol{\xi}_k\}, \{\boldsymbol{\psi}_k\}, \{\boldsymbol{\Sigma}_k\}, \ell_c)\sim \mathcal{N}_{[0,+\infty[}(a_{c}, A_c)$ \\with $A_c=\frac{1}{1+\boldsymbol{\psi}_{\ell_c}'\boldsymbol{\Sigma}_{\ell_c}^{-1}\boldsymbol{\psi}_{\ell_c}}$ and $a_c=A_c\boldsymbol{\psi}_{\ell_c}'\boldsymbol{\Sigma}_{\ell_c}^{-1}(\boldsymbol{z}_c-\boldsymbol{\xi}_{\ell_c})$
	
	\item for $k=1,\dots,K$ update $\boldsymbol{\xi}_k$, $\boldsymbol{\psi}_k$ and $\boldsymbol{\Sigma}_k$ given $G_0$, $\{\ell_{1:C}\}$ and $\{s_{1:C}\}$ from \\
	$p(\{\boldsymbol{\xi}_k\}, \{\boldsymbol{\psi}_k\}, \{\boldsymbol{\Sigma}_k\} | \{\boldsymbol{z}_{1:C}\}, \alpha, G_0, \{\ell_{1:C}\}, \{w_k\}, \{s_{1:C}\})$:
	\begin{enumerate}
		\item update $G_k$ given $\{\boldsymbol{z}_{1:C}\}$, $G_0$, $\{\ell_{1:C}\}$ and $\{s_{1:C}\}$:
		\begin{itemize}
			\item $G_0=sNiW(\boldsymbol{b}_0^\xi,\,\boldsymbol{b}_0^\psi, \boldsymbol{B}_0, \boldsymbol{\Lambda}_0, \lambda_0)$ with $\boldsymbol{b}_0=(\boldsymbol{b}_0^\xi\,'\,\boldsymbol{b}_0^\psi\,')'$ and $\boldsymbol{B}_0=diag(D_0^\xi,\,D_0^\psi)$ 
			\item $G_k=sNiW(\boldsymbol{b}_k^\xi,\,\boldsymbol{b}_k^\psi, \boldsymbol{B}_k, \boldsymbol{\Lambda}_k, \lambda_k)$ with $\boldsymbol{b}_k=(\boldsymbol{b}_k^\xi\,'\,\boldsymbol{b}_k^\psi\,')'$ 
			\item let $\boldsymbol{X}_k$ be a matrix of dimension $\text{card}(\{c|\ell_c=k\}) \times 2$:  $\boldsymbol{X}_k=(\boldsymbol{1}\ s_{c|\ell_c=k})$
			\item let $\boldsymbol{B}_k=(\boldsymbol{X}_k'\boldsymbol{X}_k + diag(\boldsymbol{D}_0)^{-1})^{-1}$
			\item $\boldsymbol{b}_k = \left(z_{c|\ell_c=k}\,\boldsymbol{X}_k + \left(\frac{1}{D_0^\xi}\boldsymbol{b}_0^\xi \ \frac{1}{D_0^\psi}\boldsymbol{b}_0^\psi\right)\right)\boldsymbol{B}_k$
			\item $\lambda_k = \lambda_0 + \text{card}(\{c|\ell_c=k\})$
			
			\item $\displaystyle\boldsymbol{\Lambda}_k = \boldsymbol{\Lambda}_0 + \sum_{c|\ell_c=k}\boldsymbol{\varepsilon}_c\boldsymbol{\varepsilon}_c' + \frac{1}{D_0^\xi}(\boldsymbol{b}_k^\xi-\boldsymbol{b}_0^\xi)(\boldsymbol{b}_k^\xi-\boldsymbol{b}_0^\xi)' + \frac{1}{D_0^\psi}(\boldsymbol{b}_k^\psi-\boldsymbol{b}_0^\psi)(\boldsymbol{b}_k^\psi-\boldsymbol{b}_0^\psi)$ \\with $\boldsymbol{\varepsilon}_c=\boldsymbol{z}_c -\boldsymbol{b}_k^\xi-s_c \boldsymbol{b}_k^\psi$
			
		\end{itemize}
		\item sample $(\boldsymbol{\xi}_k, \boldsymbol{\psi}_k, \boldsymbol{\Sigma}_k | G_k)\sim G_k$
		\begin{itemize}
			\item $((\boldsymbol{\xi}_k, \boldsymbol{\psi}_k)|\boldsymbol{\Sigma}_k, \{\ell_{1:C}\}, \{s_{1:C}\}, G_k) \sim \mathcal{N}_{2d}\left((\boldsymbol{b}_k^\xi,\boldsymbol{b}_k^\psi), \boldsymbol{B}_k\otimes\boldsymbol{\Sigma}_k\right)$
			\item $(\boldsymbol{\Sigma}_k | \{\ell_{1:C}\}, \{s_{1:C}\}, G_k) \sim \mathcal{W}^{-1}(\lambda_k, \boldsymbol{\Lambda}_k)$
		\end{itemize}
	\end{enumerate}
\end{enumerate}

\subsection{Skew $t$-distributions mixture}

Our Gibbs sampler for non parametric skew $t$-distributions mixture proceeds with each of the following updates in turn:

\begin{enumerate}
	
	\item update concentration parameter $\alpha$ given $\{\ell_{1:C}\}$ using the data augmentation technique from \cite{West1992}:\\
		$\alpha \propto p(\alpha | \{\boldsymbol{z}_{1:C}\}, G_0, \{\boldsymbol{\xi}_k\}, \{\boldsymbol{\psi}_k\}, \{\boldsymbol{\Sigma}_k\}, \{\nu_k\}, \{\ell_{1:C}\}, \{w_k\}, \{s_{1:C}\}, \{\gamma_{1:C}\}) \propto p(\alpha |\{\ell_{1:C}\})$\\
		$(\alpha,x | \{\ell_{1:C}\})\sim p(\alpha)\alpha^{K-1}(\alpha+C)x^\alpha(1-x)^{C-1}$\\
		$(x | \alpha, \{\ell_{1:C}\}) \sim \text{Beta}(\alpha+1, C)$\\
		$(\alpha | x, \{\ell_{1:C}\}) \sim \pi_x \text{Gamma}(a + K, b-\log(x)) + (1-\pi_x) \text{Gamma}(a + K-1, b-\log(x))$ with $p(\alpha)\propto \text{Gamma}(a,b)$ and $\frac{\pi_x}{1-\pi_x}=\frac{a+k-1}{C(b-\log(x))}$

	\item update $G$ given $\alpha$,  $\{\boldsymbol{\xi}_k\}$, $\{\boldsymbol{\psi}_k\}$, $\{\boldsymbol{\Sigma}_k\}$, $\{\nu_k\}$ and $G_0$ via slice sampling:\\
	$\{w_k\} , \{\ell_{1:C}\} \propto p(\{w_k\}, \{\ell_{1:C}\} | \{\boldsymbol{z}_{1:C}\}, \alpha, G_0, \{\boldsymbol{\xi}_k\}, \{\boldsymbol{\psi}_k\}, \{\boldsymbol{\Sigma}_k\}, \{\nu_k\}, \{s_{1:C}\}, \{\gamma_{1:C}\})$
	
	\begin{enumerate}
		\item sample the weights:\\
		$(w_1, \dots, w_K, w_* | \{\ell_{1:C}\}) \sim \text{Dirichlet} \left(\text{card}(\{\ell_{1:C}\}=1), \dots, \text{card}(\{\ell_{1:C}\}=K), \alpha\right)$
	
		\item for $c=1,\dots, C$: 
		$u_c\sim\text{Unif}([0, w_{\ell_c}])$
		
		\item Set $j=K$. While $\sum_{k=1}^jw_k < (1-min(u_{1:C}))$:
			\begin{itemize}
				\item set $j=j+1$
				\item sample $\pi_j\sim\text{Beta}(1,\alpha)$
				\item set $w_j =w_*\pi_j\prod_{k={K+1}}^{j-1}(1-\pi_k)$
				\item sample $(\boldsymbol{\xi}_j, \boldsymbol{\psi}_j, \boldsymbol{\Sigma}_j | G_0)\sim \text{structured-Normal-invWishart}(G_0)$
				\item sample $\nu_j\sim p(\nu_j)$
			\end{itemize}
			
		\item $K = j$
		
		\item for $c=1,\dots,C$ sample $\ell_c$ given $\{\boldsymbol{\xi}_k\}$, $\{\boldsymbol{\psi}_k\}$, $\{\boldsymbol{\Sigma}_k\}$, $\{w_k\}$ from:\\$p(\ell_c=k) \propto \mathds{1}_{\{w_{k} > u_c\}}f_{\mathcal{SN}}(\boldsymbol{z}_c, \boldsymbol{\xi}_k, \boldsymbol{\psi}_k, \boldsymbol{\Sigma}_k)$
	\end{enumerate}

		\item for $c=1,\dots,C$  update $s_c$ given $\ell_c$ , $\{\boldsymbol{\xi}_k\}$, $\{\boldsymbol{\psi}_k\}$, $\{\boldsymbol{\Sigma}_k\}$:\\
		$(s_c | \boldsymbol{z}_c, \{\boldsymbol{\xi}_k\}, \{\boldsymbol{\psi}_k\}, \{\boldsymbol{\Sigma}_k\}, \ell_c)\sim \mathcal{N}_{[0,+\infty[}(a_{c}, A_c)$ \\with $A_c=\frac{1}{1+\boldsymbol{\psi}_{\ell_c}'\boldsymbol{\Sigma}_{\ell_c}^{-1}\boldsymbol{\psi}_{\ell_c}}$ and $a_c=A_c\boldsymbol{\psi}_{\ell_c}'\boldsymbol{\Sigma}_{\ell_c}^{-1}(\boldsymbol{z}_c-\boldsymbol{\xi}_{\ell_c})$
	
	\item for $k=1,\dots,K$ update $\boldsymbol{\xi}_k$, $\boldsymbol{\psi}_k$ and $\boldsymbol{\Sigma}_k$ given $G_0$, $\{\ell_{1:C}\}$ and $\{s_{1:C}\}$	from:\\
	$p(\{\boldsymbol{\xi}_k\}, \{\boldsymbol{\psi}_k\}, \{\boldsymbol{\Sigma}_k\} | \{\boldsymbol{z}_{1:C}\}, \alpha, G_0, \{\nu_k\}, \{\ell_{1:C}\}, \{w_k\}, \{s_{1:C}\}, \{\gamma_{1:C}\})$\\$\propto p(\{\boldsymbol{\xi}_k\}, \{\boldsymbol{\psi}_k\}, \{\boldsymbol{\Sigma}_k\} | \{\ell_{1:C}\}, \{s_{1:C}\}, G_0)$:
	\begin{enumerate}
		\item update the hyper parameters of the cluster distribution given $\{\boldsymbol{z}_{1:C}\}$, $G_0$, $\{\ell_{1:C}\}$ and $\{s_{1:C}\}$:
		\begin{itemize}
			\item $G_0=sNiW(\boldsymbol{b}_0^\xi,\,\boldsymbol{b}_0^\psi, \boldsymbol{B}_0, \boldsymbol{\Lambda}_0, \lambda_0)$ with $\boldsymbol{b}_0=vec(\boldsymbol{b}_0^\xi,\,\boldsymbol{b}_0^\psi)$ and $\boldsymbol{B}_0=diag(D_0^\xi,\,D_0^\psi)$ 
			\item $G_k=sNiW(\boldsymbol{b}_k^\xi,\,\boldsymbol{b}_k^\psi, \boldsymbol{B}_k, \boldsymbol{\Lambda}_k, \lambda_k)$ with $\boldsymbol{b}_k=vec(\boldsymbol{b}_k^\xi,\,\boldsymbol{b}_k^\psi)$ 
			\item let $\boldsymbol{X}_k$ be a matrix of dimension $\text{card}(\{c|\ell_c=k\}) \times 2$:  $\boldsymbol{X}_k=(\boldsymbol{1}\ s_{c|\ell_c=k})$
			\item let $\boldsymbol{B}_k=(\boldsymbol{X}_k'\boldsymbol{X}_k + (\boldsymbol{B}_0)^{-1})^{-1}$
			\item $\boldsymbol{b}_k = \left(z_{c|\ell_c=k}\,\boldsymbol{X}_k + \left(\frac{1}{D_0^\xi}\boldsymbol{b}_0^\xi \ \frac{1}{D_0^\psi}\boldsymbol{b}_0^\psi\right)\right)\boldsymbol{B}_k$
			\item $\lambda_k = \lambda_0 + \text{card}(\{\ell_c=k\})$
			
			\item $\boldsymbol{\Lambda}_k = \boldsymbol{\Lambda}_0 + \displaystyle{\sum_{c|\ell_c=k}}\boldsymbol{\varepsilon}_c\boldsymbol{\varepsilon}_c' + \frac{1}{D_0^\xi}(\boldsymbol{b}_k^\xi-\boldsymbol{b}_0^\xi)(\boldsymbol{b}_k^\xi-\boldsymbol{b}_0^\xi)' + \frac{1}{D_0^\psi}(\boldsymbol{b}_k^\psi-\boldsymbol{b}_0^\psi)(\boldsymbol{b}_k^\psi-\boldsymbol{b}_0^\psi)$ \\with $\boldsymbol{\varepsilon}_c=\boldsymbol{z}_c -\boldsymbol{b}_k^\xi-s_c \boldsymbol{b}_k^\psi$
			
		\end{itemize}
		\item sample $(\boldsymbol{\xi}_k, \boldsymbol{\psi}_k, \boldsymbol{\Sigma}_k | \boldsymbol{b}_k, \boldsymbol{B}_k, \boldsymbol{\Lambda}_k, \lambda_k)$ from a $sNiW(\boldsymbol{b}_k, \boldsymbol{B}_k, \boldsymbol{\Lambda}_k, \lambda_k)$
		\begin{itemize}
			\item $((\boldsymbol{\xi}_k, \boldsymbol{\psi}_k)|\boldsymbol{\Sigma}_k, \{\ell_{1:C}\}, \{s_{1:C}\}, G_k) \sim \mathcal{N}_{2d}\left((\boldsymbol{b}_k^\xi,\boldsymbol{b}_k^\psi), \boldsymbol{B}_k\otimes\boldsymbol{\Sigma}_k\right)$
			\item $(\boldsymbol{\Sigma}_k | \{\ell_{1:C}\}, \{s_{1:C}\}, G_k) \sim \mathcal{W}^{-1}(\lambda_k, \boldsymbol{\Lambda}_k)$
		\end{itemize}
	\end{enumerate}
	
	\item update the degrees of freedom $\{\nu_k\}$ and the scale factors $\{\gamma_{1:C}\}$ from the random effects representation given $\{\boldsymbol{\xi}_k\}$, $\{\boldsymbol{\psi}_k\}$, $\{\boldsymbol{\Sigma}_k\}$, $\{s_{1:C}\}$ and $\{\ell_{1:C}\}$, sampling from:\\ $p(\nu_k, \{\gamma_{1:C}\} | \{\boldsymbol{\xi}_k\}, \{\boldsymbol{\psi}_k\}, \{\boldsymbol{\Sigma}_k\}, \{s_{1:C}\}, \{\ell_{1:C}\} )$
	\begin{enumerate}
		\item for $k=1,\dots,K$ update $\nu_k$, given  $\boldsymbol{\xi}_k$, $\boldsymbol{\psi}_k$, $\boldsymbol{\Sigma}_k$, $\{s_{1:C}\}$ and $\{\ell_{1:C}\}$, \text{integrating out the $\{\gamma_{1:C}\}$},  sampling from:\\
	$p(\nu_k | \{\boldsymbol{\xi}_k\}, \{\boldsymbol{\psi}_k\}, \{\boldsymbol{\Sigma}_k\}, \{\ell_{1:C}\}, \{w_k\}, \{s_{1:C}\}, \alpha, \{\gamma_{1:C}\})$\\$\propto p(\nu_k | \{\boldsymbol{\xi}_k\}, \{\boldsymbol{\psi}_k\}, \{\boldsymbol{\Sigma}_k\}, \{\ell_{1:C}\}, \{s_{1:C}\}, \{\gamma_{1:C}\})$\\$\propto p(\nu_k | \{\boldsymbol{\xi}_k\}, \{\boldsymbol{\psi}_k\}, \{\boldsymbol{\Sigma}_k\}, \{\ell_{1:C}\}, \{s_{1:C}\})$ (reducing conditioning on the $\{\gamma_{1:C}\}$)
	
	A Metropolis-Hastings step is required to sample from the above distribution. We use a uniform log random-walk proposal as proposed in \cite{Fruhwirth-Schnatter2010a}:\smallskip
	$$\log(\nu_k^{new}-1)\sim \text{Unif}([\log(\nu_k-1)-c_{\nu_k}, \log(\nu_k-1)+c_{\nu_k}]$$
	where $c_{\nu_k}$ is a fixed parameter of the algorithm (that can be tuned to improve the acceptance rate of this MH step). Acceptance probability for $\nu_k^{new}$ is as follow:\smallskip
	$$\displaystyle\min\left(1, \dfrac{p(y| \{\boldsymbol{\xi}_k\}, \{\boldsymbol{\psi}_k\}, \{\boldsymbol{\Sigma}_k\}, \nu_{-k}, \nu_k^{new}, \{\ell_{1:C}\})p(\nu_k^{new})(\nu_k^{new}-1)}{p(y| \{\boldsymbol{\xi}_k\}, \{\boldsymbol{\psi}_k\}, \{\boldsymbol{\Sigma}_k\}, \{\nu_k\}, \{\ell_{1:C}\})p(\nu_k)(\nu_k^-1)}\right)$$\smallskip
		
	\item for $c=1,\dots,C$ update $\gamma_c$ given $\{\boldsymbol{\xi}_k\}$, $\{\boldsymbol{\psi}_k\}$, $\{\boldsymbol{\Sigma}_k\}$, $\{\nu_k\}$, $s_c$ and $\ell_c$ sampling from:\\
	$\displaystyle p(\gamma_c | \{\boldsymbol{\xi}_k\}, \{\boldsymbol{\psi}_k\}, \{\boldsymbol{\Sigma}_k\}, \{\nu_k\}, s_c, \ell_c)\sim\text{Gamma}\left(\frac{\nu_{\ell_c} + d +1}{2}, \frac{\nu_{\ell_c}+\boldsymbol{z}_c^2+tr(\eta^{\phantom{'}}_c\eta_c^{'}\Sigma^{-1}_{\ell_c})}{2}\right)$\\with $\eta_c=\boldsymbol{z}_c -\boldsymbol{\xi}_{\ell_c}-s_c \boldsymbol{\psi}_{\ell_c}$
	
	\end{enumerate}
	
\end{enumerate}

\subsubsection{MH within collapsed Gibbs}

As an MH is used in the skew $t$ sampler to sample $\{\nu_k\}$, it is important to never integrate out those $\{\nu_k\}$ in the previous steps of the Partially Collapsed Gibbs sampler \citep{VanDyk2015}. Otherwise, there is no guaranty that the stationary distribution of the Markov chain remains unchanged (the correlation structure of the $\{\nu_k\}$ with the other parameters is likely to not be estimated properly). Besides, the reduced conditioning on the $\{\gamma_{1:C}\}$ does not change the stationary distribution as those marginalized out $\{\gamma_{1:C}\}$ are sampled right after the MH step from their full conditional distribution \citep{VanDyk2015}.

\section{Parameter estimation for Normal inverse-Wishart and structured Normal inverse-Wishart distributions}
\label{app:ParamEst}

\subsection{Maximum Likelihood Estimation}

\subsubsection{Maximum Likelihood estimators for Normal inverse-Wishart}

Let observations $\left(\boldsymbol{\mu}_i, \boldsymbol{\Sigma}_i\right)$ follow a Normal inverse-Wishart distribution for $i=1\dots n$:
$$\left(\boldsymbol{\mu}_i, \boldsymbol{\Sigma}_i\right)\sim NiW\left(\boldsymbol{\mu}_0, \kappa_0, \boldsymbol{\Lambda}_0, \lambda_0\right)$$

The likelihood is: 
\begin{align*}
p\left(\{\boldsymbol{\mu}_i\}_{1:n}, \{\boldsymbol{\Sigma}_i\}_{1:n} \lvert \boldsymbol{\mu}_0, \kappa_0, \boldsymbol{\Lambda}_0, \lambda_0\right) = & \prod_{i=1}^n \Bigg\{(2\pi)^{-\frac{d}{2}}|\boldsymbol{\Sigma}_i|^{-\frac{\lambda_0+d+1}{2}}\frac{2^{-\frac{\lambda_0 d}{2}}\left|\boldsymbol{\Lambda}_0\right|^{\frac{\lambda_0}{2}}}{\Gamma_d(\frac{\lambda_0}{2})} \left|\frac{1}{\kappa_0}\boldsymbol{\Sigma}_i\right|^{-\frac{1}{2}}\\ 
& \exp\left[-\frac{1}{2}tr\left(\Lambda_0\boldsymbol{\Sigma}_i^{-1}\right)-\frac{\kappa_0}{2}(\boldsymbol{\mu}_i-\boldsymbol{\mu}_0)'\boldsymbol{\Sigma}_i^{-1}(\boldsymbol{\mu}_i-\boldsymbol{\mu}_0)\right]\Bigg\}
\end{align*}
\medskip

\noindent Taking the partial derivatives of the loglikelihood with respect of the four parameters $\boldsymbol{\mu}_0, \kappa_0, \boldsymbol{\Lambda}_0, \lambda_0$ and setting each of them to zero gives the following system:
$$\begin{cases}
\displaystyle\mu_0=\sum\limits_{i=1}^{n}\boldsymbol{\mu}_i^{'}\boldsymbol{\Sigma}_i^{-1}\left(\sum_{i=1}^n \boldsymbol{\Sigma}_i^{-1}\right)^{-1}\medskip\\
\displaystyle\frac{1}{\kappa_0}=\frac{1}{nd}\sum_{i=1}^n(\boldsymbol{\mu}_i-\boldsymbol{\mu}_0)'\boldsymbol{\Sigma}_i^{-1}(\boldsymbol{\mu}_i-\boldsymbol{\mu}_0)\medskip\\
\displaystyle\boldsymbol{\Lambda}_0=n\lambda_0\left(\sum_{i=1}^n \boldsymbol{\Sigma}_i^{-1}\right)^{-1}\medskip\\
\displaystyle0=-\frac{1}{2}\sum_{i=1}^n\log\left(|\boldsymbol{\Sigma}_i|\right) - \frac{n d}{2}\log(2) + \frac{n}{2}\log\left(\left|\boldsymbol{\Lambda}_0\right|\right) -\frac{n}{2}\digamma_d\left(\frac{\lambda_0}{2}\right)\\
\end{cases}
$$
where $\displaystyle\digamma_d(x)=\frac{d}{dx}\log(\Gamma_d(x))$ is the $d$-dimensional digamma function (the derivative of the logarithm of the $d$-dimensional Gamma function).\smallskip

\noindent \textbf{NB:} The above solution are obtained using the two following identities: $\displaystyle\frac{d}{d\boldsymbol{X}}\log(|\boldsymbol{X}|)=\boldsymbol{X}^{-1}$ and $\displaystyle\frac{d}{d\boldsymbol{X}}tr(\boldsymbol{X}\boldsymbol{A})=\boldsymbol{A}'$ if $\boldsymbol{X}$ is definite-positive

\noindent Hence the MLE solutions verify:
$$\begin{cases}
\displaystyle\widehat{\boldsymbol{\mu}_0}=\sum\limits_{i=1}^{n}\boldsymbol{\mu}_i^{'}\boldsymbol{\Sigma}_i^{-1}\left(\sum_{i=1}^n \boldsymbol{\Sigma}_i^{-1}\right)^{-1}\medskip\\
\displaystyle\widehat{\kappa_0}=nd\left(\sum_{i=1}^n(\boldsymbol{\mu}_i-\widehat{\boldsymbol{\mu}_0})'\boldsymbol{\Sigma}_i^{-1}(\boldsymbol{\mu}_i-\widehat{\boldsymbol{\mu}_0})\right)^{-1}\medskip\\
\displaystyle \digamma_d\left(\frac{\widehat{\lambda_0}}{2}\right)=-\frac{1}{n}\sum_{i=1}^n\log\left(|\boldsymbol{\Sigma}_i|\right) +  d\log\left(\frac{n\widehat{\lambda_0}}{2}\right) - \log\left(\left|\sum_{i=1}^n \boldsymbol{\Sigma}_i^{-1}\right|\right)\medskip\\
\displaystyle\widehat{\boldsymbol{\Lambda}_0}=n\widehat{\lambda_0}\left(\sum_{i=1}^n \boldsymbol{\Sigma}_i^{-1}\right)^{-1}\\
\end{cases}
$$
under the constraint $\widehat{\lambda_0}>d+1$ (in which case there should a unique solution $\widehat{\lambda_0}$).

\subsubsection{Maximum Likelihood estimators for structured Normal inverse-Wishart}

Let observations $\left(\boldsymbol{\xi}_i, \boldsymbol{\psi}_i, \boldsymbol{\Sigma}_i\right)$ follow a structured Normal inverse-Wishart distribution ($sNiW$) for $i=1\dots n$:
$$\left(\boldsymbol{\xi}_i, \boldsymbol{\psi}_i, \boldsymbol{\Sigma}_i\right)\sim sNiW\left(\boldsymbol{\xi}_0, \boldsymbol{\psi}_0, \boldsymbol{B}_0, \boldsymbol{\Lambda}_0, \lambda_0\right)$$

\noindent The likelihood is: 
\begin{align*}
p\left(\{\boldsymbol{\xi}_i\}_{1:n}, \{\boldsymbol{\psi}_i\}_{1:n}, \{\boldsymbol{\Sigma}_i\}_{1:n} \lvert \boldsymbol{\mu}_0, \boldsymbol{B}_0, \boldsymbol{\Lambda}_0, \lambda_0\right) = & \prod_{i=1}^n \Bigg\{(2\pi)^{-\frac{d}{2}}|\boldsymbol{\Sigma}_i|^{-\frac{\lambda_0+d+1}{2}}\frac{2^{-\frac{\lambda_0 d}{2}}\left|\boldsymbol{\Lambda}_0\right|^{\frac{\lambda_0}{2}}}{\Gamma_d(\frac{\lambda_0}{2})} \left|\boldsymbol{B}_0^{-1}\otimes\boldsymbol{\Sigma}_i\right|^{-\frac{1}{2}}\\ 
& \exp\Big[-\frac{1}{2}tr\left(\boldsymbol{\Lambda}_0\boldsymbol{\Sigma}_i^{-1}\right)\\
&-\frac{1}{2}(\boldsymbol{\mu}_i-\boldsymbol{\mu}_0)' \left(\boldsymbol{B}_0\otimes\boldsymbol{\Sigma}_i^{-1}\right)(\boldsymbol{\mu}_i-\boldsymbol{\mu}_0)\Big]\Bigg\}
\end{align*}
where $\boldsymbol{\mu}_i = (\boldsymbol{\xi}_i'\ \boldsymbol{\psi}_i')'$ and $\boldsymbol{\mu}_0 = (\boldsymbol{\xi}_0'\ \boldsymbol{\psi}_0')'$\medskip

\noindent Taking the partial derivatives of the loglikelihood with respect of the four parameters $\mu_0, B_0, \Lambda_0, \lambda_0$ and setting each of them to zero gives the following system:
$$\begin{cases}
\displaystyle\boldsymbol{\mu}_0=\sum\limits_{i=1}^{n}\boldsymbol{\mu}_i^{'}\boldsymbol{\Sigma}_i^{-1}\left(\sum_{i=1}^n \boldsymbol{\Sigma}_i^{-1}\right)^{-1}\medskip\\
\displaystyle 0=\sum_{i=1}^n\left(\frac{d}{d\boldsymbol{B_0}}\left(\log\left(\left|\boldsymbol{B_0}^{-1}\otimes\boldsymbol{\Sigma}_i\right|\right)\right) + \frac{d}{d\boldsymbol{B_0}}\left((\boldsymbol{\mu}_i-\boldsymbol{\mu}_0)'\left(\boldsymbol{B_0}\otimes\boldsymbol{\Sigma}_i^{-1}\right)(\boldsymbol{\mu}_i-\boldsymbol{\mu}_0)\right)\right)\medskip\\
\displaystyle\boldsymbol{\Lambda}_0=n\lambda_0\left(\sum_{i=1}^n \boldsymbol{\Sigma}_i^{-1}\right)^{-1}\medskip\\
\displaystyle0=-\frac{1}{2}\sum_{i=1}^n\log\left(|\boldsymbol{\Sigma}_i|\right) - \frac{n d}{2}\log(2) + \frac{n}{2}\log\left(\left|\boldsymbol{\Lambda}_0\right|\right) -\frac{n}{2}\digamma_d\left(\frac{\lambda_0}{2}\right)\\
\end{cases}
$$
where $\displaystyle\digamma_d(x)=\frac{d}{dx}\log(\Gamma_d(x))$ is the digamma function (the derivative of the logarithm of the Gamma function).

\begin{align*}
&\sum_{i=1}^n\left(\frac{d}{d\boldsymbol{B_0}}\left(\log\left(\left|\boldsymbol{B_0}^{-1}\otimes\boldsymbol{\Sigma}_i\right|\right)\right) + \frac{d}{d\boldsymbol{B_0}}\left((\boldsymbol{\mu}_i-\boldsymbol{\mu}_0)'\left(\boldsymbol{B_0}\otimes\boldsymbol{\Sigma}_i^{-1}\right)(\boldsymbol{\mu}_i-\boldsymbol{\mu}_0)\right)\right)\\
=&\displaystyle\sum_{i=1}^n\frac{d}{d\boldsymbol{B_0}}\left(\log\left(\left|\boldsymbol{B_0}\right|^{-d}\left|\boldsymbol{\Sigma}_i\right|^2\right)\right)  + \sum_{i=1}^n\frac{d}{d\boldsymbol{B_0}}\left((\boldsymbol{\mu}_i-\boldsymbol{\mu}_0)' \left(\boldsymbol{B}_0 \otimes\boldsymbol{\Sigma}_i^{-1}\right)(\boldsymbol{\mu}_i-\boldsymbol{\mu}_0)\right)\medskip\\
=&\displaystyle - nd\frac{d}{d\boldsymbol{B_0}}\left(\log\left(\left|\boldsymbol{B_0}\right|\right)\right)  + \sum_{i=1}^n\frac{d}{d\boldsymbol{B_0}}\left(tr\left((\boldsymbol{\mu}_i-\boldsymbol{\mu}_0)' \left(\boldsymbol{B}_0 \otimes\boldsymbol{\Sigma}_i^{-1}\right)(\boldsymbol{\mu}_i-\boldsymbol{\mu}_0)\right)\right)\medskip\\
=&\displaystyle -nd\boldsymbol{B_0}^{-1} + \sum_{i=1}^n\left(\begin{array}{cc}\boldsymbol{\xi}_i -\boldsymbol{\xi}_0 & \boldsymbol{\psi}_i -\boldsymbol{\psi}_0\end{array}\right)' \left(\boldsymbol{\Sigma}_i^{-1}\right)\left(\begin{array}{cc}\boldsymbol{\xi}_i -\boldsymbol{\xi}_0 & \boldsymbol{\psi}_i -\boldsymbol{\psi}_0\end{array}\right)\medskip\\
=&\displaystyle -nd\boldsymbol{B_0}^{-1} + \sum_{i=1}^n\left(\begin{array}{c}\boldsymbol{\xi}_i'-\boldsymbol{\xi}_0'\\ \boldsymbol{\psi}_i' -\boldsymbol{\psi}_0'\end{array}\right) \left(\boldsymbol{\Sigma}_i^{-1}\right)\left(\begin{array}{cc}\boldsymbol{\xi}_i -\boldsymbol{\xi}_0 & \boldsymbol{\psi}_i -\boldsymbol{\psi}_0\end{array}\right)\medskip\\
\end{align*}

\noindent So if the above expression is zero, we get:
\begin{align*}
&\displaystyle \widehat{\boldsymbol{B}_0} = nd \left(\sum_{i=1}^n \left(\begin{array}{c}\boldsymbol{\xi}_i'-\boldsymbol{\xi}_0'\\ \boldsymbol{\psi}_i' -\boldsymbol{\psi}_0'\end{array}\right) \left(\boldsymbol{\Sigma}_i^{-1}\right)\left(\begin{array}{cc}\boldsymbol{\xi}_i -\boldsymbol{\xi}_0 & \boldsymbol{\psi}_i -\boldsymbol{\psi}_0\end{array}\right)\right)^{-1}\medskip\\
\end{align*}

So MLE solution for $sNiW$ are:\\
	$$\begin{cases}
\displaystyle\widehat{\boldsymbol{\mu}_0}=\sum\limits_{i=1}^{n}\boldsymbol{\mu}_i^{'}\boldsymbol{\Sigma}_i^{-1}\left(\sum_{i=1}^n \boldsymbol{\Sigma}_i^{-1}\right)^{-1}\medskip\\
\displaystyle \widehat{\boldsymbol{B}_0} = nd \left(\sum_{i=1}^n \left(\begin{array}{c}\boldsymbol{\xi}_i'-\boldsymbol{\xi}_0'\\ \boldsymbol{\psi}_i' -\boldsymbol{\psi}_0'\end{array}\right) \left(\boldsymbol{\Sigma}_i^{-1}\right)\left(\begin{array}{cc}\boldsymbol{\xi}_i -\boldsymbol{\xi}_0 & \boldsymbol{\psi}_i -\boldsymbol{\psi}_0\end{array}\right)\right)^{-1}\medskip\\
\displaystyle \digamma_d\left(\frac{\widehat{\lambda_0}}{2}\right)=-\frac{1}{n}\sum_{i=1}^n\log\left(|\boldsymbol{\Sigma}_i|\right) +  d\log\left(\frac{n\widehat{\lambda_0}}{2}\right) - \log\left(\left|\sum_{i=1}^n \boldsymbol{\Sigma}_i^{-1}\right|\right)\medskip\\
\displaystyle\widehat{\boldsymbol{\Lambda}_0}=n\widehat{\lambda_0}\left(\sum_{i=1}^n \boldsymbol{\Sigma}_i^{-1}\right)^{-1}\\
\end{cases}
$$

\subsection{Expectation-Maximization algorithms (MLE \& $MAP$)}
\subsubsection{MLE estimation via an E-M algorithm}
\label{ss:MLE_EM}

The latent variables used in the EM algorithm for estimating a finite mixture model over the MCMC draws for the parameters $\boldsymbol{\xi}_i$, $\boldsymbol{\psi}_i$ and $\boldsymbol{\Sigma}_i$ are the allocation variables $\ell_i$, with $i=1..N$ the number of (MCMC) observations. An (MCMC) observation is then $\boldsymbol{x}_i=(\boldsymbol{\xi}_i, \boldsymbol{\psi}_i, \boldsymbol{\Sigma}_i)$. Let $K$ be the number of components in the mixture model:

$$p(\boldsymbol{x}_i|K, \boldsymbol{\theta}_{\{1:K\}})=\sum_{k=1}^K \pi_k f_{\boldsymbol{\theta}_{\ell_i}}(\boldsymbol{x}_i|\ell_i, \boldsymbol{\theta}_{\{1:K\}}) \qquad \text{for } i=1\dots N $$

where $f_{\boldsymbol{\theta}_k}$ is the parametric density function of a cluster: a $sNiW$ density function with parameters $\boldsymbol{\theta}_k=\left(\boldsymbol{\xi}_k, \boldsymbol{\psi}_k, \boldsymbol{B}_k, \boldsymbol{\Lambda}_k, \lambda_k\right)$.\smallskip

At iteration $t$, the EM algorithm maximizes $Q\left(\boldsymbol{\theta}_{\{1:K\}}\left|\boldsymbol{\theta}_{\{1:K\}}^{(t-1)}\right.\right)$ for $\boldsymbol{\theta}_{\{1:K\}}$ with:
	\begin{align*}Q\left(\boldsymbol{\theta}_{\{1:K\}}\left|\boldsymbol{\theta}_{\{1:K\}}^{(t-1)}\right.\right) &= {E}\left[\left.\log\left(p(\boldsymbol{x}_{\{1:n\}}, \ell_{\{1:N\}}|K, \boldsymbol{\theta}_{\{1:K\}})\right)\right|\boldsymbol{\theta}_{\{1:K\}}^{(t-1)}\right]\\
	&=\sum_{\ell_{\{1:N\}}}\log\left(p(\boldsymbol{x}_{\{1:n\}}, \ell_{\{1:N\}}|K, \boldsymbol{\theta}_{\{1:K\}}^{(t-1)})\right)\\
	&=\sum_{k=1}^K\sum_{i=1}^n r_{ik}^{(t-1)}\log(\pi_{k}) + \sum_{k=1}^K\sum_{i=1}^n r_{ik}^{(t-1)}\log\left(p(\boldsymbol{x}_i|K, \boldsymbol{\theta}_{\{1:K\}})\right)\\
	&= \sum_{k=1}^K\sum_{i=1}^n \Bigg[ r_{ik}^{(t-1)}\log(\pi_{k}) -\frac{\lambda_k + d + 1}{2} r_{ik}^{(t-1)}\log(|\boldsymbol{\Sigma}_i|)\\
	&\phantom{=}-\frac{\lambda_k d r_{ik}^{(t-1)}}{2}\log(2) -r_{ik}^{(t-1)}\log(\Gamma_d(\frac{\lambda_k}{2})) + \frac{r_{ik}^{(t-1)}\lambda_k}{2}\log(|\boldsymbol{\Lambda}_k|)\\
	&\phantom{=} - \frac{r_{ik}^{(t-1)}}{2}\log(|\boldsymbol{B}_k^{-1}\otimes\boldsymbol{\Sigma}_i|) - \frac{r_{ik}^{(t-1)}}{2}tr(\boldsymbol{\Lambda}_k\boldsymbol{\Sigma}_i^{-1})\\
	&\phantom{=} -\frac{r_{ik}^{(t-1)}}{2}(\boldsymbol{\mu}_i-\boldsymbol{\mu}_k)'(\boldsymbol{B}_k\otimes\boldsymbol{\Sigma}_i^{-1})(\boldsymbol{\mu}_i-\boldsymbol{\mu}_k)\Bigg]
	\end{align*}
	
	\noindent with $\displaystyle r_{ik}^{(t)}=p(\ell_i=k|\boldsymbol{x}_i, \boldsymbol{\theta}_{\{1:k\}}^{(t)})=\frac{\pi_k f_{\boldsymbol{\theta}_{k}^{(t)}}(\boldsymbol{x}_i)}{\sum_{j=1}^K \pi_jf_{\boldsymbol{\theta}_{j}^{(t)}}(\boldsymbol{x}_i)}$

\begin{enumerate}
	\item \textbf{Initialization}
	
	$\boldsymbol{\theta}^{(0)}_k$ is initialized randomly ($\pi_k$ are initialized at $1/K$)

	\item \textbf{E step at iteration $t$}
	
	Compute the membership weights $r_{ik}^{(t-1)}$ for each observation $i=1\dots N$ for each cluster $k=1\dots K$:
	
	$$r_{ik}^{(t-1)}=p\left(\ell_i=k\left|\boldsymbol{x}_i, \boldsymbol{\theta}_{\{1:k\}}^{(t-1)}\right.\right)=\frac{\pi_k f_{\boldsymbol{\theta}_{k}^{(t-1)}}(\boldsymbol{x}_i)}{\sum_{j=1}^K \pi_jf_{\boldsymbol{\theta}_{j}^{(t-1)}}(\boldsymbol{x}_i)}$$

	\item \textbf{M step at iteration $t$}
	
	Update the parameters:
	\begin{itemize}	
		\item	$\boldsymbol{\theta}^{(t)}_k$ are updated with their weighted Maximum Likelihood Estimators for each k:
			$$\begin{cases}
\displaystyle\widehat{\boldsymbol{\mu}_k}=\sum\limits_{i=1}^{n}r_{ik}^{(t-1)}\boldsymbol{\mu}_i^{'}\boldsymbol{\Sigma}_i^{-1}\left(\sum_{i=1}^n r_{ik}^{(t-1)}\boldsymbol{\Sigma}_i^{-1}\right)^{-1}\medskip\\
\displaystyle \widehat{\boldsymbol{B}_k} = N_kd \left(\sum_{i=1}^{n} r_{ik}^{(t-1)} \left(\begin{array}{c}\boldsymbol{\xi}_i'-\widehat{\boldsymbol{\xi}_k}'\\ \boldsymbol{\psi}_i' -\widehat{\boldsymbol{\psi}_k}'\end{array}\right) \left(\boldsymbol{\Sigma}_i^{-1}\right)\left(\begin{array}{cc}\boldsymbol{\xi}_i-\widehat{\boldsymbol{\xi}_k} & \boldsymbol{\psi}_i -\widehat{\boldsymbol{\psi}_k}\end{array}\right)\right)^{-1}\medskip\\
\displaystyle \digamma_d\left(\frac{\widehat{\lambda_k}}{2}\right)=-\frac{1}{N_k}\sum_{i=1}^n r_{ik}^{(t-1)}\log\left(|\boldsymbol{\Sigma}_i|\right)+  d\log\left(\frac{N_k\widehat{\lambda_k}}{2}\right) - \log\left(\left|\sum_{i=1}^n r_{ik}^{(t-1)}\boldsymbol{\Sigma}_i^{-1}\right|\right)\medskip\\
\displaystyle\widehat{\boldsymbol{\Lambda}_k}={N_k}\widehat{\lambda_k}\left(\sum_{i=1}^n r_{ik}^{(t-1)}\boldsymbol{\Sigma}_i^{-1}\right)^{-1}\\
\end{cases}
$$
	 	\item	$\pi^{(t)}_k$ are updated with $N_k/n$, $N_k=\sum_{i=1}^n r_{ik}$
	\end{itemize}
	
	\item \textbf{Repeat 2. and 3. until convergence}
	
	Convergence is reached when the incomplete log-likelihood $l^{(t)}$ is unchanged between two consecutive iterations $t$ and $t+1$ of the 2. and 3. steps:
	$$l^{(t)} = \log\left(p(\boldsymbol{x}_{\{1:N\}}|K, \boldsymbol{\theta}_{\{1:K\}}^{(t)})\right)= \sum_{i=1}^n \log\left(\sum_{k=1}^K\pi_k p(\boldsymbol{x}_i|K, \boldsymbol{\theta}_{\{1:K\}}^{(t)})\right)$$  
	
\end{enumerate}

\subsubsection{$MAP$ estimation via E-M algorithm}
\label{ss:MAP_EM}

In order to avoid degenrate covariance matrices (for instance when $K$ is set to too many clusters in the EM algorithm), it can be useful to replace MLE estimation with Maximum A Posteriori ($MAP$) estimations \citep{Fraley2007}.

To perform a $MAP$ estimation instead of a MLE estimation as in section \ref{ss:MLE_EM}, the E-step of the algorithm is unchanged, but the M-step now maximizes the following $Q$ function:

 	\begin{align*}Q\left(\boldsymbol{\theta}_{\{1:K\}}\left|\boldsymbol{\theta}_{\{1:K\}}^{(t-1)}\right.\right) &= {E}\left[\left.\log\left(p(\boldsymbol{\theta}_{\{1:K\}})p(\boldsymbol{x}_{\{1:n\}}, \ell_{\{1:n\}}|K, \boldsymbol{\theta}_{\{1:K\}})\right)\right|\boldsymbol{\theta}_{\{1:K\}}^{(t-1)}\right]\\
	&=\sum_{\ell_{\{1:N\}}}\left(\log\left(p(\boldsymbol{x}_{\{1:n\}}, \ell_{\{1:n\}}|K, \boldsymbol{\theta}_{\{1:K\}}^{(t-1)})\right)\right) + \log(p(\boldsymbol{\theta}_{\{1:K\}}))\\
	&=\log(p(\boldsymbol{\theta}_{\{1:K\}})) + \sum_{k=1}^K\sum_{i=1}^n r_{ik}^{(t-1)}\log(\pi_{k}) + \sum_{k=1}^K\sum_{i=1}^N r_{ik}^{(t-1)}\log\left(p(\boldsymbol{x}_i|K, \boldsymbol{\theta}_{\{1:K\}})\right)
	\end{align*}

We use the following priors :

\begin{itemize}

\item a Dirichlet prior over the cluster weigths $\pi_{\{1:K\}}$ with all parameters equal to the same $\alpha$ ( if $\alpha=1$, then this is equivalent to a uniform prior over the $K-1$ simplex):
$$(\pi_1,\dots,\pi_K)\sim Dir(\alpha)$$

\end{itemize}

\noindent And for each $k$:

\begin{itemize}

\item a Normal-Wishart empirical bayes prior on $\left(\boldsymbol{\mu}_k, \boldsymbol{B}_k\right)$:

$$\left(\boldsymbol{\mu}_k, \boldsymbol{B}_k\right)\sim\mathcal{NW}\left(\boldsymbol{m}, \kappa_0, \boldsymbol{C}, 4\right)$$
\begin{align*}
	\boldsymbol{\mu}_k | \boldsymbol{m}, \kappa_0,  \boldsymbol{B}_k, \boldsymbol{\Sigma}_{\{1:n\}} \sim& \mathcal{N}\left(\boldsymbol{m}, \frac{1}{\kappa_0}\left( \boldsymbol{B}_k\otimes\frac{1}{n}\sum_{i=1}^n\boldsymbol{\Sigma}_i^{-1}\right)^{-1}\right)\\
	\boldsymbol{B}_k | \boldsymbol{C} \sim& \mathcal{W}\left(\boldsymbol{C}, 4\right)
\end{align*}

with $\boldsymbol{m}=\overline{\boldsymbol{\mu}}_{\{1:n\}}$, $\displaystyle \boldsymbol{C}=100 \boldsymbol{I}_2$ and $\displaystyle \boldsymbol{L}=(\boldsymbol{S}^{(\xi)} + \boldsymbol{S}^{(\psi)})/2$ (where $\boldsymbol{S}^{(\xi)}=diag(var(\boldsymbol{\xi}_{\{1:n\}}))$ and $\boldsymbol{S}^{(\psi)}=diag(var(\boldsymbol{\psi}_{\{1:n\}}))$) and $\kappa_0 =0.01$ for instance. The harmonic mean is used as an empirical bayes prior for the bloc variance matrix.

\noindent One can also specify a vague prior on $\boldsymbol{\boldsymbol{\mu}}_k$: $ \boldsymbol{\mu}_k\sim \mathcal{U}^{2d}_{]-\infty, +\infty[}$ (which simplifies the $\boldsymbol{\xi}$ and $\boldsymbol{\psi}$ $MAP$ estimators, as long as no cluster has an exactly null $0$ contribution $N_k$)

\item a Wishart priors on $\boldsymbol{\Lambda}_k$:
$$\boldsymbol{\Lambda}_k\sim \mathcal{W}\left(\boldsymbol{L}, d+2\right)$$
with $\displaystyle \boldsymbol{L}=(\boldsymbol{S}^{(\xi)} + \boldsymbol{S}^{(\psi)})/2$ (where $\boldsymbol{S}^{(\xi)}=diag(var(\boldsymbol{\xi}_{\{1:n\}}))$ and $\boldsymbol{S}^{(\psi)}=diag(var(\boldsymbol{\psi}_{\{1:n\}}))$) 

\item an Exponential prior on $\lambda_k$ under the constraint that $\lambda_k \geq d+1$ :

$$\lambda_k - (d+1) \sim Exp(1)$$

\end{itemize}

\begin{align*}Q(\boldsymbol{\theta}|\boldsymbol{\theta}^{(t-1)}) =& \sum_{k=1}^K \Bigg[ -\frac{1}{2}\log\left(\left|\boldsymbol{B}_k\otimes\left(\sum_{i=1}^n\boldsymbol{\Sigma}_i^{-1}\right)^{-1}\right|\right)-\frac{\kappa_0}{2n}(\boldsymbol{\mu}_k -\boldsymbol{m})' \left(\boldsymbol{B}_k\otimes\sum_{i=1}^n\boldsymbol{\Sigma}_i^{-1}\right)(\boldsymbol{\mu}_k - \boldsymbol{m})\\
& + \frac{1}{2}\log(|\boldsymbol{B}_k|) -\frac{1}{2}tr\left(\boldsymbol{C}^{-1}\boldsymbol{B}_k\right) + \frac{1}{2}\log(|\boldsymbol{\Lambda}_k|) -\frac{1}{2}tr\left(\boldsymbol{L}^{-1}\boldsymbol{\Lambda}_k\right) - \lambda_k \Bigg]\\
& +\sum_{k=1}^K\sum_{i=1}^n \Bigg[ r_{ik}^{(t-1)}\log(\pi_{k}) -\frac{\lambda_k + d + 1}{2} r_{ik}^{(t-1)}\log(|\boldsymbol{\Sigma}_i|)\\
	&-\frac{\lambda_k d r_{ik}^{(t-1)}}{2}\log(2) -r_{ik}^{(t-1)}\log(\Gamma_d(\frac{\lambda_k}{2})) + \frac{r_{ik}^{(t-1)}\lambda_k}{2}\log(|\boldsymbol{\Lambda}_k|)\\
	& - \frac{r_{ik}^{(t-1)}}{2}\log(|\boldsymbol{B}_k\otimes\boldsymbol{\Sigma}_i|) - \frac{r_{ik}^{(t-1)}}{2}tr(\boldsymbol{\Lambda}_k\boldsymbol{\Sigma}_i^{-1})\\
	&-\frac{r_{ik}^{(t-1)}}{2}(\boldsymbol{\mu}_i-\boldsymbol{\mu}_k)'(\boldsymbol{B}_k\otimes\boldsymbol{\Sigma}_i^{-1})(\boldsymbol{\mu}_i-\boldsymbol{\mu}_k)\Bigg] + constant
\end{align*}

\begin{align*}
%\frac{dQ(\theta|\theta^{(t-1)})}{d\pi_k} =&\medskip\\
%\frac{dQ(\theta|\theta^{(t-1)})}{d\xi_k} =& \frac{d}{d\xi_k}\Bigg(-\frac{1}{2}(\xi_k - m^{(\xi)})' {S^{(\psi)}}^{-1}(\xi_k - m^{(\xi)})\\
%&-\sum_{i=1}^n \bigg[\frac{r_{ik}}{2}(\boldsymbol{\mu}_i-\boldsymbol{\mu}_k)'(\boldsymbol{B}_k\otimes\boldsymbol{\Sigma}_i^{-1})(\boldsymbol{\mu}_i-\boldsymbol{\mu}_k)\bigg]\Bigg)\medskip\\
%\frac{dQ(\theta|\theta^{(t-1)})}{d\psi_k} =& \frac{d}{d\psi_k}\Bigg(-\frac{1}{2}(\psi_k - m^{(\psi)})' {S^{(\psi)}}^{-1}(\psi_k - m^{(\psi)})\\
%&-\sum_{i=1}^n \bigg[\frac{r_{ik}}{2}(\boldsymbol{\mu}_i-\boldsymbol{\mu}_k)'(\boldsymbol{B}_k\otimes\boldsymbol{\Sigma}_i^{-1})(\boldsymbol{\mu}_i-\boldsymbol{\mu}_k)\bigg]\Bigg)\medskip\\
\frac{dQ(\boldsymbol{\theta}|\boldsymbol{\theta}^{(t-1)})}{d\lambda_k} =& -\frac{N_k}{2}\digamma_d\left(\frac{\widehat{\lambda_k}}{2}\right)-\frac{1}{2}\sum_{i=1}^n r_{ik}^{(t-1)}\log\left(|\boldsymbol{\Sigma}_i|\right)+  \frac{N_kd}{2}\log\left(\frac{N_k\widehat{\lambda_k}}{2}\right)\medskip\\
&- \frac{N_k}{2}\log\left(\left|\sum_{i=1}^n r_{ik}^{(t-1)}\boldsymbol{\Sigma}_i^{-1}\right|\right)-1\medskip\\
\frac{dQ(\boldsymbol{\theta}|\boldsymbol{\theta}^{(t-1)})}{d\boldsymbol{B}_k} =&  \frac{d}{2}\boldsymbol{B}_k + \frac{N_{k}d}{2}\boldsymbol{B}_k -  \sum_{i=1}^n \frac{r_{ik}^{(t-1)}}{2}\left(\begin{array}{c}\boldsymbol{\xi}_i'-\boldsymbol{\xi}_k'\\ \boldsymbol{\psi}_i' -\boldsymbol{\psi}_k'\end{array}\right) \left(\boldsymbol{\Sigma}_i^{-1}\right)\left(\begin{array}{cc}\boldsymbol{\xi}_i-\boldsymbol{\xi}_k & \boldsymbol{\psi}_i -\boldsymbol{\psi}_k\end{array}\right)\\
& -\frac{\kappa_0}{2n} \left(\begin{array}{c}\boldsymbol{\xi}_k'-\boldsymbol{m}^{(\xi)}\,'\\ \boldsymbol{\psi}_k' -\boldsymbol{m}^{(\psi)}\,'\end{array}\right) \sum_{i=1}^n\boldsymbol{\Sigma}_i^{-1} \left(\begin{array}{cc}\boldsymbol{\xi}_k-\boldsymbol{m}^{(\xi)} & \boldsymbol{\psi}_k -\boldsymbol{m}^{(\psi)}\end{array}\right) + \frac{1}{2}\boldsymbol{B}_k-\frac{1}{2}\boldsymbol{C}^{-1}
\medskip\\
\frac{dQ(\boldsymbol{\theta}|\boldsymbol{\theta}^{(t-1)})}{d\boldsymbol{\Lambda}_k} =& \frac{N_k\lambda_k}{2}\boldsymbol{\Lambda}_k^{-1} -\frac{1}{2}\sum_{i=1}^n r_{ik}^{(t-1)}\boldsymbol{\Sigma}_i^{-1} + \frac{1}{2}\boldsymbol{\Lambda}_k^{-1}-\frac{1}{2}\boldsymbol{L}^{-1}\\
\end{align*}

The $MAP$ estimators of $\boldsymbol{\theta}_k |\boldsymbol{\theta}_k^{(t-1)}$ are thus:

$$\begin{cases}
\displaystyle\widehat{\pi_k}^{MAP}= \frac{N_k + \alpha -1}{n + K(\alpha - 1)}\medskip\\
\displaystyle\widehat{\boldsymbol{\mu}_k}^{MAP}=\sum\limits_{i=1}^{n}\boldsymbol{m}^{'}\boldsymbol{\Sigma}_i^{-1}+ r_{ik}^{(t-1)}\boldsymbol{\mu}_i^{'}\boldsymbol{\Sigma}_i^{-1}\left(\sum_{i=1}^n \frac{\kappa_0}{n}\boldsymbol{\Sigma}_i^{-1} +r_{ik}^{(t-1)}\boldsymbol{\Sigma}_i^{-1}\right)^{-1}\medskip\\
\displaystyle \widehat{\boldsymbol{B}_k}^{MAP} = (N_kd + d+1) \Bigg[\boldsymbol{C}^{-1} + \sum_{i=1}^{n} r_{ik}^{(t-1)} \left(\begin{array}{c}\boldsymbol{\xi}_i'-\widehat{\boldsymbol{\xi}_k}^{MAP}\,'\\ \boldsymbol{\psi}_i' -\widehat{\boldsymbol{\psi}_k}^{MAP}\,'\end{array}\right)\left(\boldsymbol{\Sigma}_i^{-1}\right)\left(\begin{array}{cc}\boldsymbol{\xi}_i-\widehat{\boldsymbol{\xi}_k}^{MAP} & \boldsymbol{\psi}_i -\widehat{\boldsymbol{\psi}_k}^{MAP}\end{array}\right)\\
+\displaystyle \frac{\kappa_0}{n}\left(\begin{array}{c}\boldsymbol{\xi}_k'-\boldsymbol{m}^{(\xi)}\,'\\ \boldsymbol{\psi}_k' -\boldsymbol{m}^{(\psi)}\,'\end{array}\right) \sum_{i=1}^n\boldsymbol{\Sigma}_i^{-1} \left(\begin{array}{cc}\boldsymbol{\xi}_k-\boldsymbol{m}^{(\xi)} & \boldsymbol{\psi}_k -\boldsymbol{m}^{(\psi)}\end{array}\right)\Bigg]^{-1}\medskip\\
\displaystyle 0 = N_k\digamma_d\left(\frac{\widehat{\lambda_k}^{MAP}}{2}\right) + \sum_{i=1}^n r_{ik}^{(t-1)}\log\left(|\boldsymbol{\Sigma}_i|\right) -  N_k d\log\left(\frac{N_k\widehat{\lambda_k}^{MAP}}{2}\right) + N_k \log\left(\left|\sum_{i=1}^n r_{ik}\boldsymbol{\Sigma}_i^{-1}\right|\right) + 2\medskip\\
\displaystyle\widehat{\boldsymbol{\Lambda}_k}^{MAP}=\left({N_k}\widehat{\lambda_k}^{MAP} + 1\right)\left(\boldsymbol{L}^{-1} +\sum_{i=1}^n r_{ik}^{(t-1)}\boldsymbol{\Sigma}_i^{-1}\right)^{-1}\\
\end{cases}
$$
with $N_k=\sum_{i=1}^n r_{ik}^{(t-1)}$

\begin{enumerate}
	\item \textbf{Initialization}
	
		$\boldsymbol{\theta}_k^{(0)}$ are initialized randomly ($\pi_k$ are initialized at $1/K$)

	\item \textbf{E step}
	
	Compute the membership weights $r_{ik}^{(t-1)}$ for each observation $i=1\dots N$ for each cluster $k=1\dots K$:

	\item \textbf{M step}
	
	Update the parameters:
	\begin{itemize}	
		\item	$\boldsymbol{\theta}_k$ are updated with their MAP estimation for each k:
			$$\begin{cases}
\displaystyle \pi_k^{(t)} =\frac{N_k + \alpha -1}{n + K(\alpha - 1)}\medskip\\
\displaystyle \boldsymbol{\mu}_k^{(t)}=\sum\limits_{i=1}^{n}\boldsymbol{m}^{'}\boldsymbol{\Sigma}_i^{-1}+ r_{ik}^{(t-1)}\boldsymbol{\mu}_i^{'}\boldsymbol{\Sigma}_i^{-1}\left(\sum_{i=1}^n \frac{\kappa_0}{n}\boldsymbol{\Sigma}_i^{-1} +r_{ik}^{(t-1)}\boldsymbol{\Sigma}_i^{-1}\right)^{-1}\medskip\\
\displaystyle \boldsymbol{B}_k^{(t)} = (N_kd + d + 1) \Bigg[\boldsymbol{C}^{-1} + \sum_{i=1}^{n} r_{ik}^{(t-1)} \left(\begin{array}{c}\boldsymbol{\xi}_i'-\boldsymbol{\xi}_0'\\ \boldsymbol{\psi}_i' - \boldsymbol{\psi}_0'\end{array}\right) \left(\boldsymbol{\Sigma}_i^{-1}\right)\left(\begin{array}{cc}\boldsymbol{\xi}_i -\boldsymbol{\xi}_0 & \boldsymbol{\psi}_i -\boldsymbol{\psi}_0\end{array}\right)\medskip\\
+\displaystyle \frac{\kappa_0}{n}\left(\begin{array}{c}\boldsymbol{\xi}_k^{(t)}\,'-\boldsymbol{m}^{(\xi)}\,'\\ \boldsymbol{\psi}_k^{(t)}\,' -\boldsymbol{m}^{(\psi)}\,'\end{array}\right) \sum_{i=1}^n\boldsymbol{\Sigma}_i^{-1} \left(\begin{array}{cc}\boldsymbol{\xi}_k^{(t)}-\boldsymbol{m}^{(\xi)} & \boldsymbol{\psi}_k^{(t)} - \boldsymbol{m}^{(\psi)}\end{array}\right)\Bigg]^{-1}\medskip\\
\displaystyle \digamma_d\left(\frac{\lambda_k^{(t)}}{2}\right)=-\frac{1}{N_k}\sum_{i=1}^n r_{ik}^{(t-1)}\log\left(|\boldsymbol{\Sigma}_i|\right)+  d\log\left(\frac{N_k \lambda_k^{(t)}}{2}\right) - \log\left(\left|\sum_{i=1}^n r_{ik}^{(t-1)}\boldsymbol{\Sigma}_i^{-1}\right|\right)\medskip\\
\displaystyle \Lambda_0^{(t)}=({N_k}\lambda_k^{(t)} +1 )\left(\boldsymbol{L}^{-1} + \sum_{i=1}^n r_{ik}^{(t-1)}\boldsymbol{\Sigma}_i^{-1}\right)^{-1}\\
\end{cases}
$$
with $N_k=\sum_{i=1}^n r_{ik}^{(t-1)}$
	\end{itemize}

	\item \textbf{Repeat 2. and 3. until convergence}
	
		Convergence is reached when the incomplete log-likelihood $l^{(t)}$ is unchanged between two consecutive iterations $t$ and $t+1$ of the 2. and 3. steps:
	$$l^{(t)} = \log\left(p(\boldsymbol{x}_{\{1:N\}}|K, \boldsymbol{\theta}_{\{1:K\}}^{(t)})\right)= \sum_{i=1}^n \log\left(\sum_{k=1}^K\pi_k p(\boldsymbol{x}_i|K, \boldsymbol{\theta}_{\{1:K\}}^{(t)})\right)$$  
	
\end{enumerate}

\section{Limited $\mathcal{F}$-measure}
\label{app:limitedF}

First let's start from a reference partition $G=\{g_1, \dots, g_m\}$ and an estimated partition $H=\{h_1, \dots, h_n\}$.
In order to compute a $\mathcal{F}$-measure limited to the clusters that have less than $p$ observations, we need to define two subartition $G^{(p)}$ and $H^{(p)}$ respectively. Let's denote $g_{q'}$ the clusters from the reference partition $G$ that have less than $p$ observations: $\{g_{q'}\} = \left\{g_q \, \big\lvert\,|g_q| < p\right\}$. Now let's consider all the estimated clusters that each contains at least one observation included in this subpartition. This gives the estimated limited partition $H^{(p)}=\{h^{(p)}_{r}\}=\left\{h_r \, \big\lvert \, \exists\, c \in h_r\cap \{g_{q'}\} \right\}$. Finally, let's consider the reference limited partition for the observations included in $H^{p}$:  $G^{(p)}$ is the reference partition induced by $\{\ell_c \,\lvert\, c\in H^{(p)}\}$. The limited $\mathcal{F}$-measure is then defined as follows:

$$\mathcal{F}_{lim}(H, G, p) = \mathcal{F}_{tot}(H^{(p)}, G^{(p)})=\frac{1}{\sum_{g\in G^{(p)}} |g|}\sum_{g\in G^{(p)}}^m |g|\,\underset{h\in H^{(p)}}{\operatorname{max}}\,\mathcal{F}(h, g)$$

\begin{figure}[!h]
	\centerline{\includegraphics[width=0.95\textwidth]{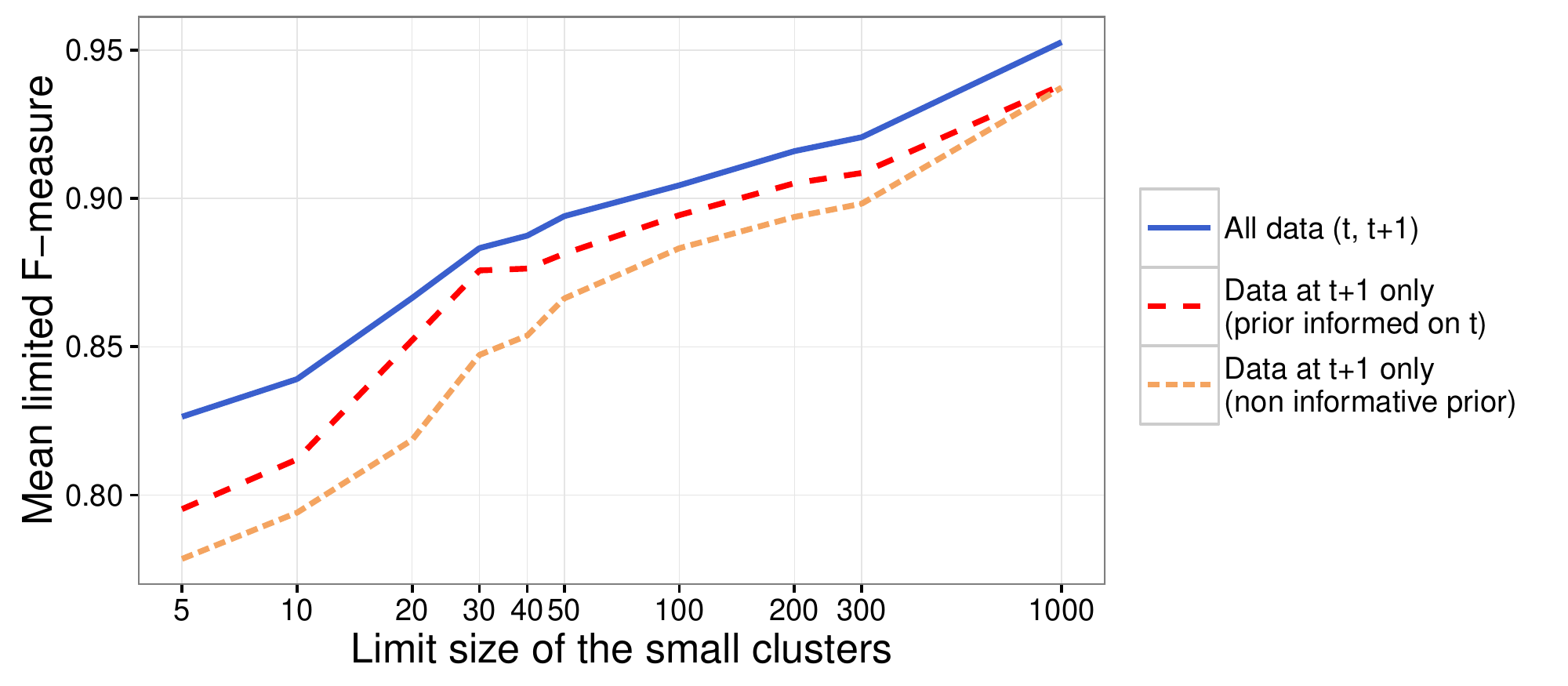}}
	\caption{Mean limited $\mathcal{F}$-measure according to the limit size of rare populations, over 300 simulations}
	\label{fig:FLimited}
\end{figure}

Figure \ref{fig:FLimited} displays the mean of this limited $\mathcal{F}$-measure for several different limit maximum size for small clusters. Thus it seems that the use of an informative prior in the sequential strategy always improves the clustering accuracy for small sized clusters.

\section{flowMeans applied to the DALIA-1 trial}
\label{app:FM}

\begin{figure}[!h]
	\centerline{\includegraphics[width=0.95\textwidth]{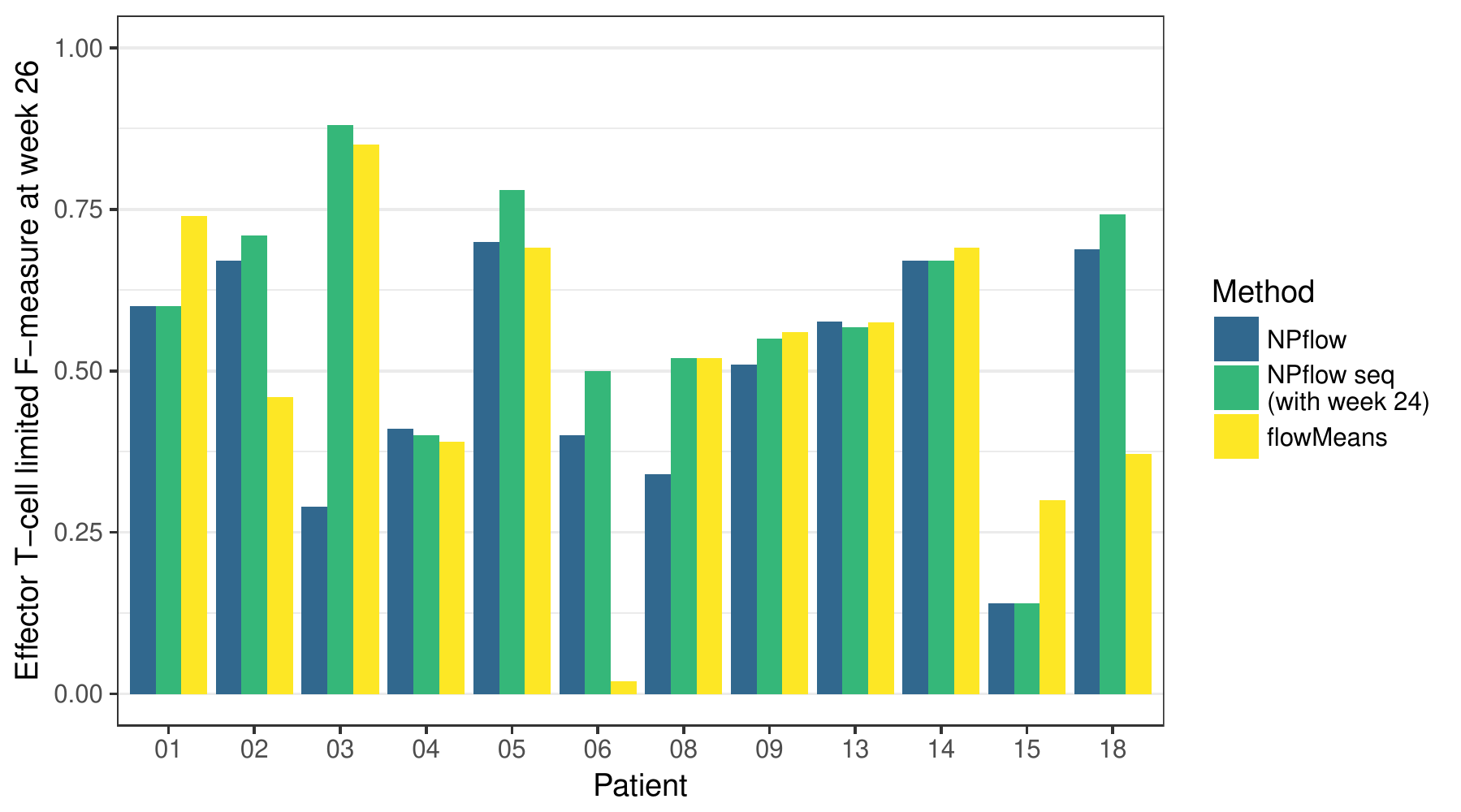}}
	\caption{Limited $\mathcal{F}$-measures for the effector T-cell population from the DALIA-1 trial two weeks after HAART interruption for NPflow with or without the sequential strategy and for flowMeans, compared to manual gating.}
	\label{fig:FmeasDalia_all}
\end{figure}

\begin{figure}[!h]
	\centerline{\includegraphics[width=0.95\textwidth]{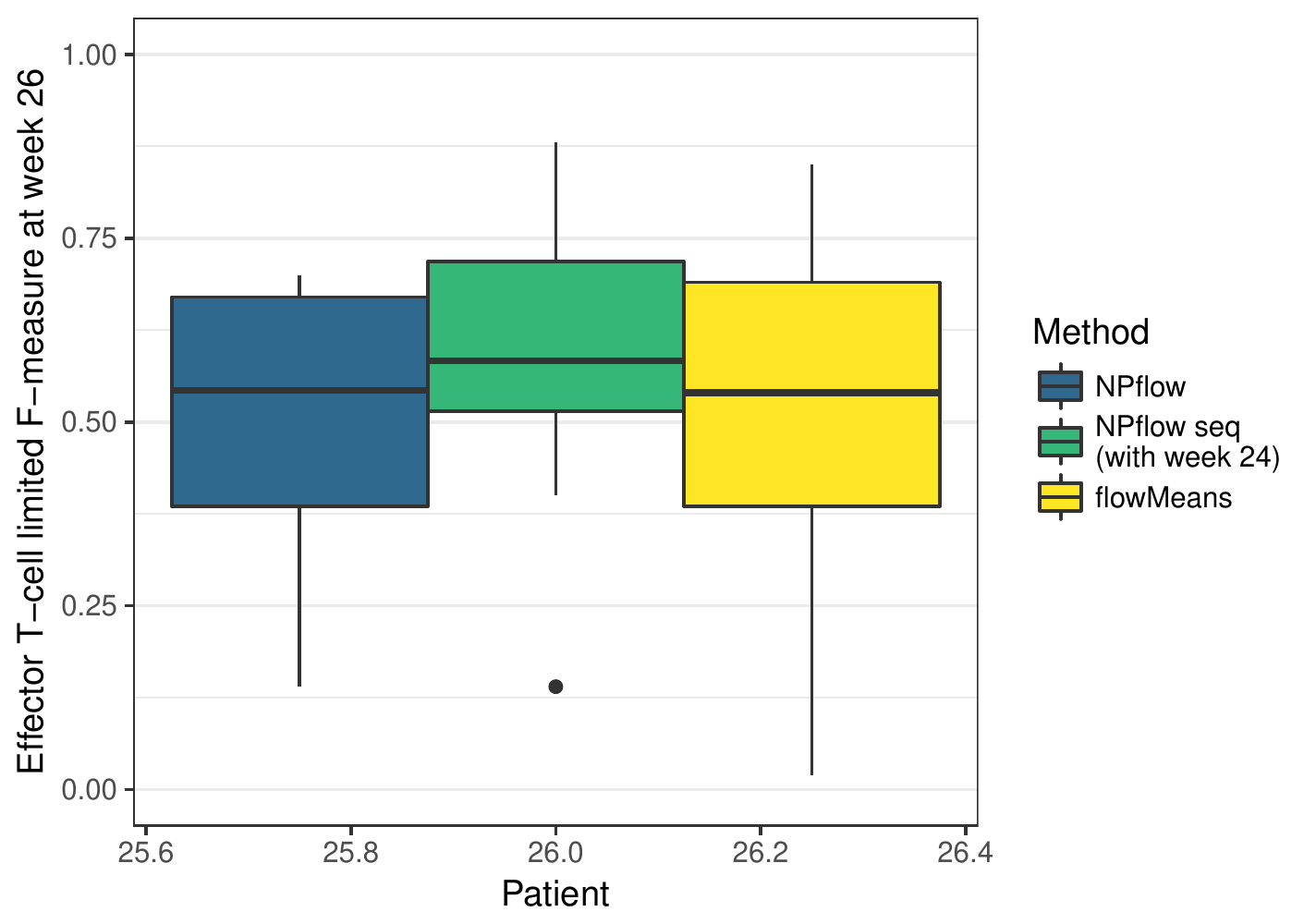}}
	\caption{Boxplots of the limited $\mathcal{F}$-measures for the effector CD4+ T-cell population from the DALIA-1 trial two weeks after HAART interruption for NPflow with or without the sequential strategy and for flowMeans, compared to manual gating.}
	\label{fig:FmeasDalia_box}
\end{figure}

\begin{figure}[!h]
	\centerline{\includegraphics[width=0.8\textwidth]{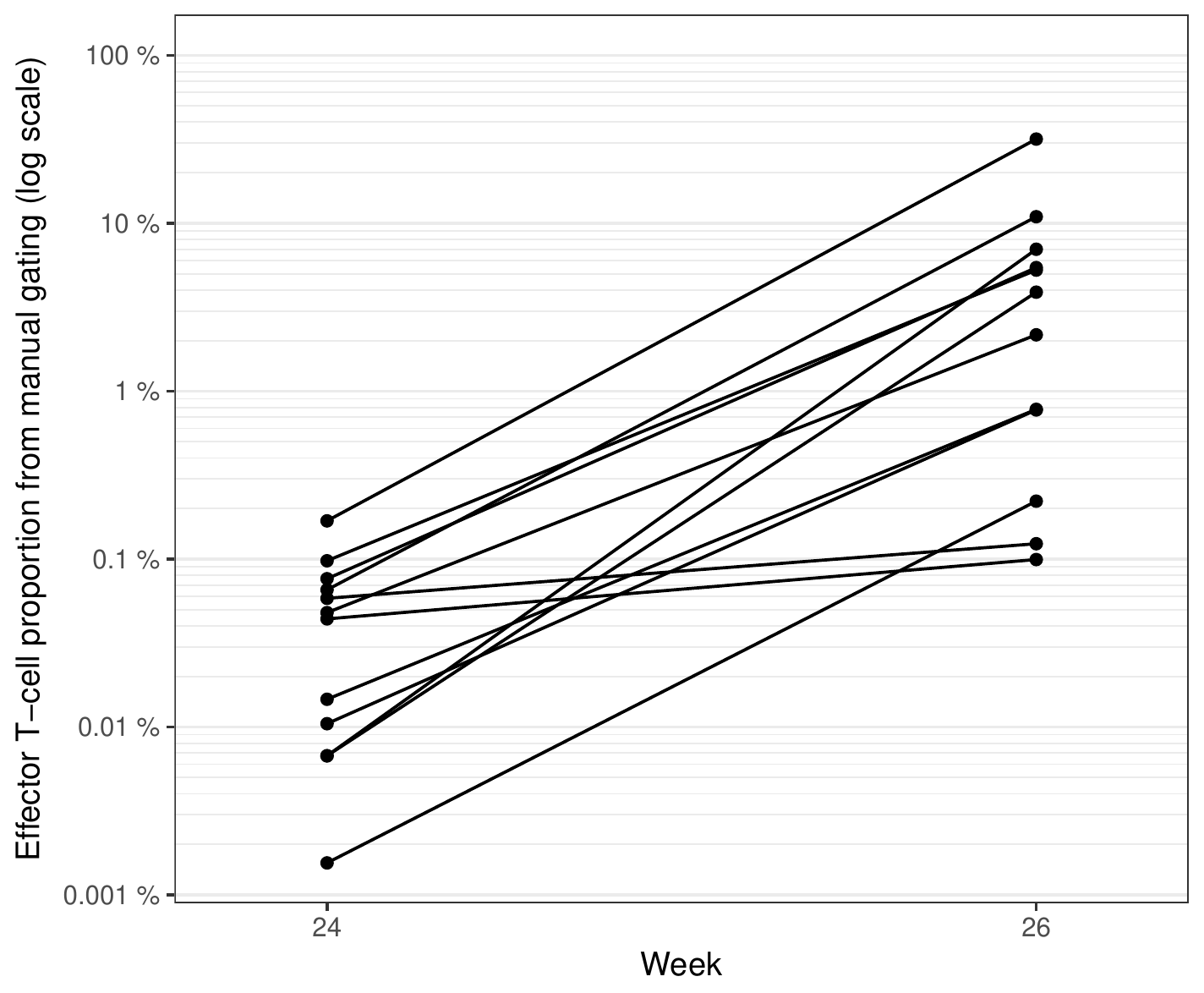}}
	\caption{Paired proportions of effector CD4+ T-cells in the DALIA-1 trial before and after HAART interruption from manual gating.}
	\label{fig:PropDalia}
\end{figure}

Here we provide additional representation of the results from flowMeans applied to the DALIA-1 trial and compared to NPflow for the effector CD4+ T-cell population. Overall the results of flowMeans are comparable to those of NPflow without the sequential posterior approximation strategy, as can be seen from Figures \ref{fig:FmeasDalia_all}, \ref{fig:FmeasDalia_box} and \ref{fig:PropDalia}, while the sequential strategy outperforms both.

\end{document}